\documentclass{paper}
\usepackage{booktabs}
\usepackage{amsmath}
\usepackage{subcaption}
\usepackage[table]{xcolor}
\usepackage[T1]{fontenc}
\usepackage{fontsetup}
\usepackage{cite}
\usepackage{amssymb}
\usepackage[pagebackref=true,colorlinks,linkcolor=blue,citecolor=blue,urlcolor=blue,bookmarks=false,pdfpagemode=UseNone]{hyperref}
\usepackage{graphicx}

\title{Perceptually Aligned Evaluation of Style Transfer}

\author[Y. Deng et al.]
{
\parbox{\textwidth}{
\centering
Yang Deng$^{1}$,
Eleftherios Ioannou$^{2}$,
David Mould$^{3}$,
Steve Maddock$^{4}$,
Paul L. Rosin$^{1}$,
and Yu-Kun Lai$^{1}$
}
\\
\parbox{\textwidth}{
\centering
$^{1}$Cardiff University, UK\\
$^{2}$The Cyprus Institute, Cyprus\\
$^{3}$Carleton University, Canada\\
$^{4}$University of Sheffield, UK
}
}

\begin{document}

\maketitle
\begin{abstract}

Style transfer lacks a reliable evaluation standard: ground truth is inherently ill-defined, and existing automatic metrics often fail to reflect human preference. This paper introduces \textbf{ASTRA (Assessment of Style TRansfer Algorithms)}, an approach for automatic evaluation of style transfer algorithms; it contains two components, \textbf{ASTRA-Data} and \textbf{ASTRA-Score}. ASTRA-Data consists of a benchmark image set of content and style references, a collection of style transfer results generated on the benchmark set, and user study data capturing human judgements through a two-stage pairwise comparison protocol. From these annotations, we derive ranking-based ground truth for content preservation, style fidelity, and overall preference. Based on ASTRA-Data, we construct ASTRA-Score, a learnt evaluator that predicts preference-aligned scores from content-style-stylization image triplets, enabling automatic and scalable evaluation of new models applied to the benchmark set. Experimental results demonstrate that ASTRA-Score achieves substantially higher correlation with human rankings compared to prior metrics. Overall, ASTRA establishes a robust mechanism for standardised evaluation of style transfer methods.


\end{abstract}

\begin{figure*}[t]
    \centering
    \includegraphics[width=\linewidth]{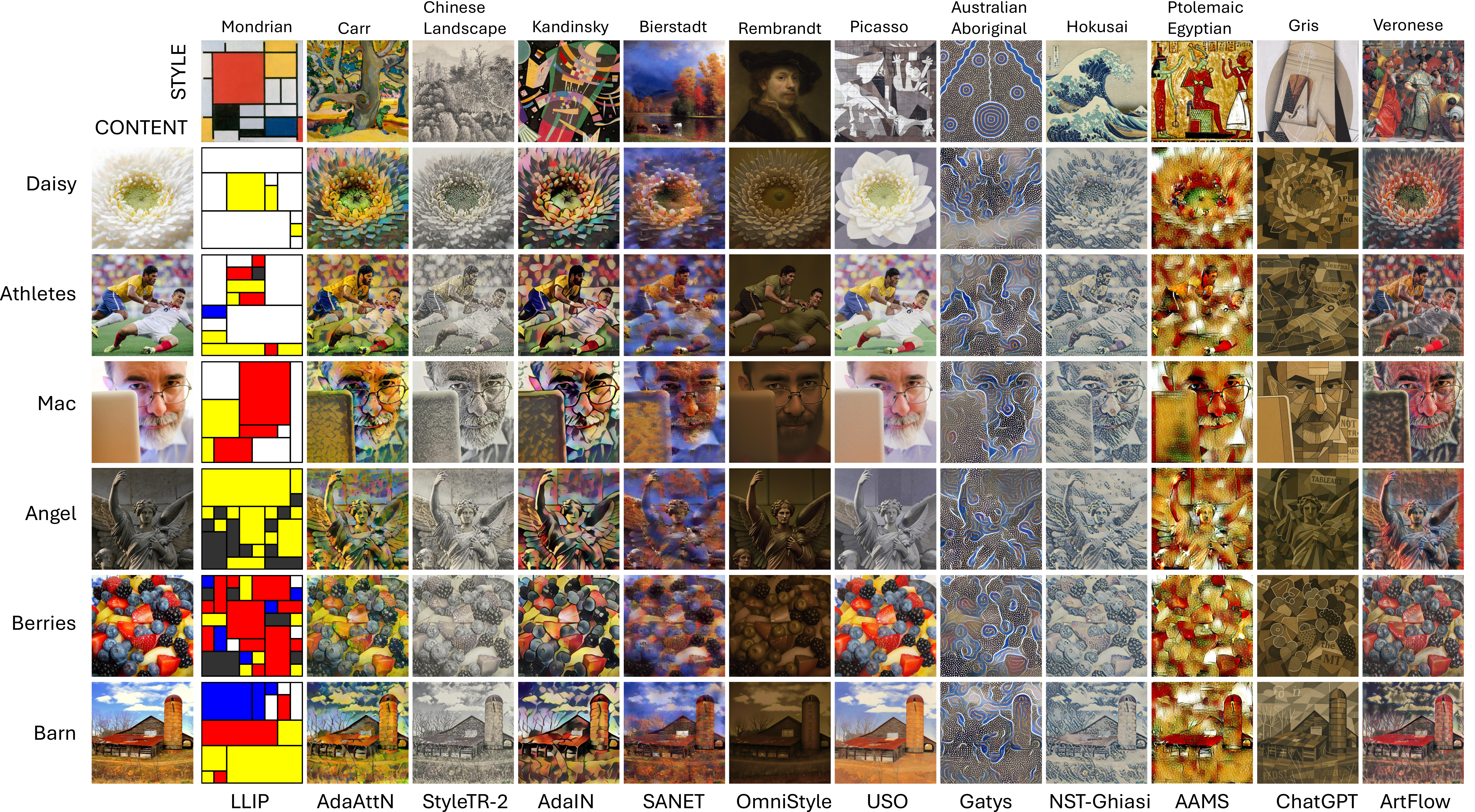}
    \caption{
    Examples of selected images produced by different algorithms.
    }

    \label{fig:examples_of_selection}
\end{figure*}
\section{Introduction}
Style transfer aims to generate an image that preserves the content of one image while adopting the style of another~\cite{jacobs2001image}.
The field has evolved from low-level image processing~\cite{rosin2010towards,lai2012non,rosin2013non,rosin2015non,rosin2017benchmarking} to modern neural style transfer (NST) methods~\cite{gatys2016image,liu2021adaattn,huang2017arbitrary,park2019arbitrary,deng2022stytr2} and has enabled a wide range of applications, including artistic image generation~\cite{ghiasi2017exploring, wang2025omnistyle2}, photo editing~\cite{huang2017arbitrary}, semantic segmentation~\cite{zhao2022source}, and data augmentation for downstream vision tasks~\cite{jackson2019style}.

Evaluating stylisation quality is fundamentally challenging~\cite{rosin2022nprportrait}, yet
reliable evaluation is vital to advancing style transfer research.
Style transfer lacks well‑defined ground truth; in some cases, there are multiple perceptually valid outputs for a single content–style pair,
while in others, it is not clear what if any output would qualify.
Consequently, evaluation is intrinsically ill‑posed~\cite{rosin2022nprportrait,mould2017developing}. Although metrics have progressed from low‑level similarity measures to high‑level representations~\cite{gatys2016image,heusel2017gans,zhang2018unreasonable,wright2022artfid}, pretrained vision and vision‑language models~\cite{dosovitskiy2020image,caron2021emerging,bai2023qwen} still treat many legitimate stylistic transformations as errors. Artistic styles often involve structural abstraction or reconfiguration~\cite{kotovenko2019content}, and in such cases, existing metrics are not reliable indicators. User studies remain the de facto standard~\cite{lei2025stylestudio,gao2025styleshot} despite their cost, limited scalability, and poor reproducibility~\cite{bylinskii2023towards}.

Evaluation is further constrained by weakly standardised test data. Early work relied on a few handpicked examples~\cite{gatys2016image}, while later studies drew from general‑purpose datasets
that were not intended for stylisation assessment. More systematic benchmarks have been proposed~\cite{mould2017developing,rosin2017benchmarking,rosin2022nprportrait,chen2022quality,ruta2024neat}, but typically provide image collections without assessment protocols. In summary, the field still lacks standardised, perceptually aligned frameworks for evaluating artistic style transfer.

This paper advances style transfer evaluation by presenting ASTRA, \textit{Assessment of Style TRansfer Algorithms}.
ASTRA contains both a benchmark dataset (ASTRA-Data, a curated dataset of style and content images) and an automated evaluation protocol (ASTRA-Score, a trained network that can estimate style transfer performance on the benchmark set). This combined benchmark and automated assessment constitutes a flexible tool for evaluating style transfer algorithms: practitioners can apply their proposed process to all combinations of style and content images in the benchmark, and use ASTRA-Score to generate numerical evaluations for each result image. These individual scores can either be aggregated or examined independently to discern particular strengths and weaknesses with respect to the benchmark coverage.

To construct this framework, we assembled a diverse collection of painterly style images capturing a representative range of historical practices and visual elements, alongside a carefully selected set of content images. By combining these with a group of influential and varied style transfer algorithms, we produced a comprehensive set of style-content-algorithm triples, yielding output result images for evaluation.

These result images provided the raw material for a large-scale user study using a two-alternative-forced-choice design. Subjects evaluated pairs based on three criteria: (i) which result better captured the style; (ii) which better preserved the content; and (iii) which was overall a more satisfactory style transfer outcome. We deliberately avoided asking which result was more appealing overall, as general aesthetics are largely orthogonal to style transfer fidelity and can be confounded by preferred colors, content, or external visual elements. We refer to this full collection (style and content images, output images, and pairwise human judgments) as ASTRA-Data.

Using rank centrality to derive global rankings from these human judgments, we trained ASTRA-Score to predict stylisation effectiveness given a style image, content image, and transfer result. A leave-one-out experimental study indicates that ASTRA-Score achieves substantially greater alignment with human judgement than past automated methods such as
ArtFID~\cite{wright2022artfid}, SRQE~\cite{chen2022quality}, and Gram Loss~\cite{gatys2016image}.

This paper contains two main contributions, corresponding to ASTRA's core components:
\begin{itemize}
\item \textbf{ASTRA-Data}, the collection of style images, content images, style transfer results, and associated human judgements, establishing a rigorous benchmark for style transfer evaluation.
\item \textbf{ASTRA-Score}, a network trained on ASTRA-Data to predict style transfer effectiveness on the benchmark image set, allowing reliable automated assessment of performance on the benchmark.
\end{itemize}

\section{Related Work}
Evaluation in style transfer remains an open problem. The field lacks a  unified approach that jointly defines evaluation data, human judgement, and quantitative metrics~\cite{ioannou2024evaluation}. Existing work typically combines manually selected test images, small-scale user studies, and proxy-based automatic metrics. These components are often designed independently, with limited reproducibility and inconsistent comparisons across methods~\cite{mould2017developing,zhou2024comprehensive}.
\subsection{Evaluation Dataset for Style Transfer.}
Some researchers evaluated results on a small set of manually chosen content–style pairs~\cite{gatys2016image}. Later studies drew from standard datasets, often sampling content images from MS COCO~\cite{lin2014microsoft} and artworks from WikiArt~\cite{mohammad2018wikiart}, with additional datasets such as KTH~\cite{karayev2013recognizing}, BBST4M~\cite{ruta2024neat}, and Behance~\cite{wilber2017bam} used to increase diversity. However, these datasets were primarily designed for general visual understanding rather than stylisation evaluation. In practice, they are used with ad-hoc sampling strategies, leading to unsystematic exploration of  content–style combinations, poor coverage of stylistic diversity, and a lack of consistency across studies. These limitations have motivated the development of dedicated evaluation datasets, such as curated NPRgeneral~\cite{mould2017developing} and human-annotated datasets like AST-IQAD~\cite{chen2022quality},
While AST-IQAD~\cite{chen2022quality} introduces human-annotated evaluation for stylisation quality and explicitly constructs controlled content–style pairs, its protocol is still based on a fixed set of pre-selected content–style pairs rather than a cross content-style evaluation design that independently probes content and style factors.
Its subjective comparisons remain confined to this fixed pool of generated outputs, making it difficult to disentangle pair-specific difficulty from general method performance. In addition,
the final objective evaluator is constructed from sparse handcrafted feature similarity rather than learned directly from human judgements.

\subsection{Evaluation Methods}
Early work primarily relied on qualitative side-by-side comparisons of stylised outputs generated from selected content–style pairs~\cite{gatys2016image}. While intuitive, such comparisons are subjective and are unstable with respect to the examples provided. User studies have been widely adopted to assess perceptual quality~\cite{mould2014authorial}, typically measuring preferences related to content preservation, style resemblance, and overall quality. Such studies are expensive and difficult to scale. Various automatic metrics have been introduced to provide complementary quantitative signals. Content preservation is often measured using perceptual similarity metrics such as LPIPS~\cite{zhang2018unreasonable} or feature similarity derived from pretrained vision models including CLIP~\cite{radford2021learning}, Vision Transformers~\cite{dosovitskiy2020image}, and DINO~\cite{caron2021emerging}. Distribution-based measures such as FID~\cite{heusel2017gans} and ArtFID~\cite{wright2022artfid} have also been adapted for stylisation evaluation. Additional metrics target stylisation quality or perceptual fidelity, including CFSD~\cite{chung2024style}, SRQE~\cite{chen2022quality}; some recent evaluation~\cite{wang2025omnistyle2} uses large vision–language models such as Qwen~\cite{bai2023qwen}. Another possibility is learned evaluators that decompose stylisation quality into perceptual components such as aesthetic quality, content preservation, and style resemblance~\cite{somepalli2024measuring,chen2024assessing}. However, these models are typically trained on proxy signals or limited annotated datasets, and their alignment with human judgement remains limited.
\begin{figure}[t]
\centering
    \begin{subfigure}{\linewidth}
        \centering
        \includegraphics[width=\linewidth]{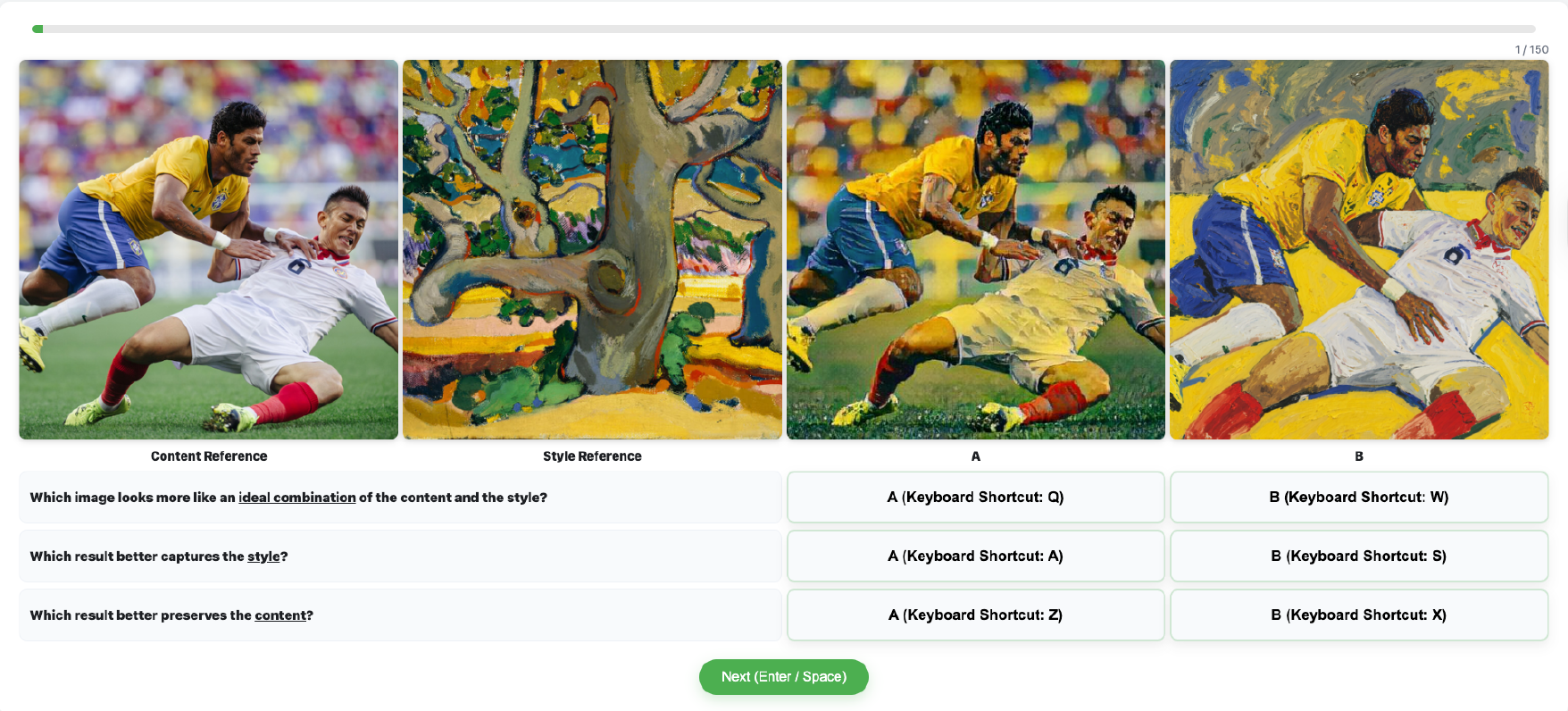}
        \caption{Stage 1 evaluation interface.}

    \end{subfigure}
    \begin{subfigure}{\linewidth}
        \centering
        \includegraphics[width=\linewidth]{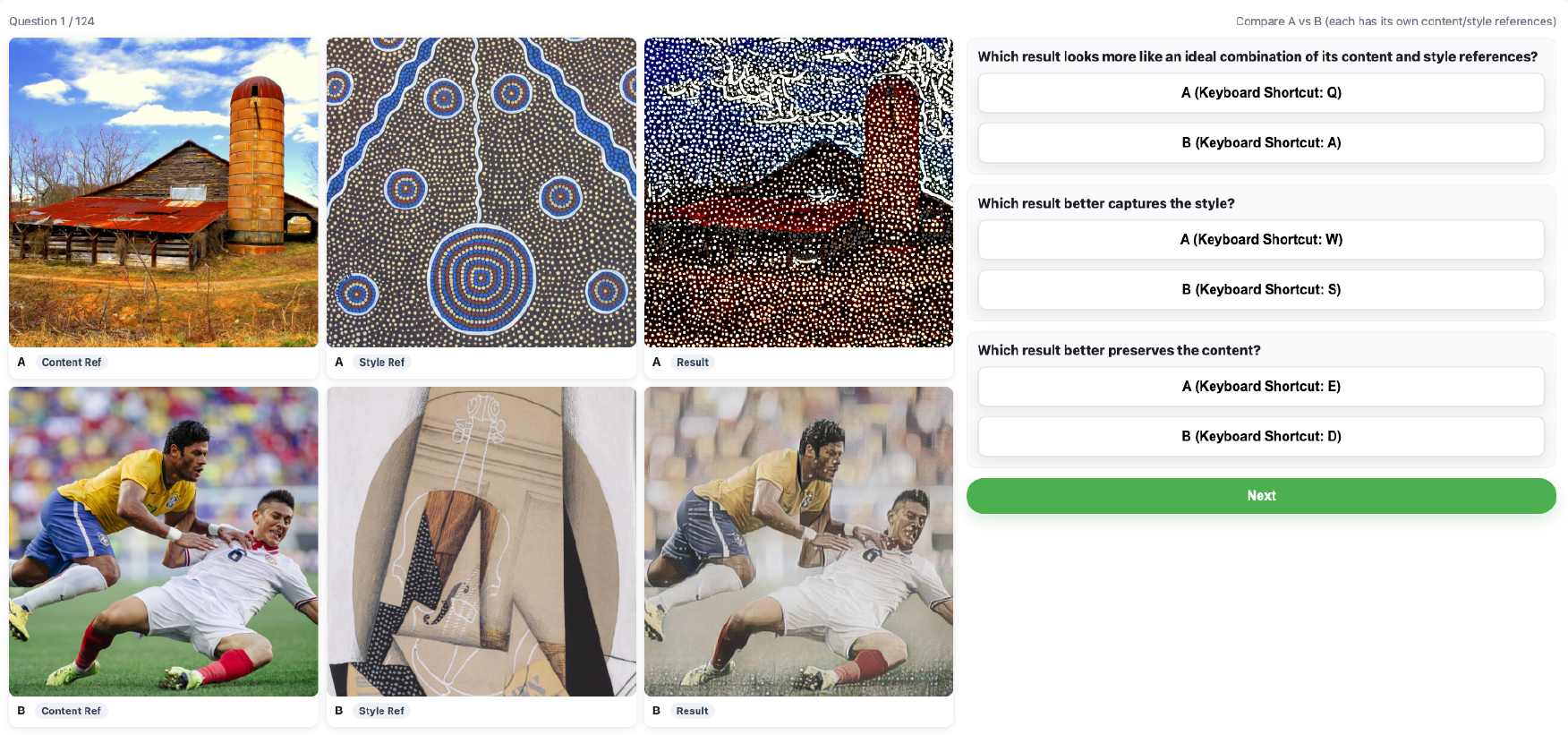}
        \caption{Stage 2 evaluation interface.}

    \end{subfigure}
    \caption{
        Examples of the two evaluation interfaces used in our user study.
        (\textbf{Top}) In Stage~1, participants compare two stylised results generated from the same content-style pair by different methods, together with the corresponding content and style references.
        (\textbf{Bottom}) In Stage~2, participants compare two stylised results that come from different content-style pairs.
        }

    \label{fig:userstudy_interface}
\end{figure}
\section{Construction of ASTRA-Data}
In this section, we describe the construction of \textbf{ASTRA-Data}, including the selection of
style images in Section~\ref{sec:style references selection},
content images in Section~\ref{sec:content references selection}, and representative stylisation methods in Section~\ref{sec:method selection}. We selected
12 style references,
6 content images,
and 12 representative methods, resulting in a total of 864 stylised outputs covering all content–style–method combinations.
This dataset is small enough to permit human evaluations of a substantial fraction of possible pairs, from which global rankings can be computed.

\subsection{Style Image Selection}
\label{sec:style references selection}
A vast array of image styles are possible. In order to balance completeness of coverage with diversity of styles across a small number of images, we opted to use only paintings, and exclude many other worthwhile art media and styles. That said, we took an expansive view in considering the eligibility of a given image, and our final set includes images such as Hokusai's \textit{Great Wave} which is a woodcut print.

We designed our image set around the three dimensions of content, colourfulness, and detail.
\emph{Content} refers to the subject matter being depicted. \emph{Colourfulness} refers to both the range of colours present in the image and the vividness of individual colours. \emph{Detail} refers to the presence of intentional fine-scale structure in the depiction of the subject matter, as opposed to more simplified or abstracted depictions that deliberately omit detail.
Organising our image search along these dimensions helped us systematically choose an image set with a broad range and distributed sampling of painterly characteristics.

One might question why we chose to include content as a dimension of style when we have a separate set of content images. A few reasons explain this. Historical painting genres and traditions are often based on content (e.g., portraits, landscapes), and elements and motifs used in style creation can be particular to the content being depicted. On a more technical note, many style transfer methods make use of spatial statistics (e.g., summarised in Gram matrices), and varying the image content helps obtain a broader sampling of styles.
We chose three categories of content:
\emph{people}, encompassing both portraits and
images where humans occupy significant portions of the canvas; \emph{nature}, including landscapes and other natural scenes;
and \emph{artificial}, meaning images that are primarily about constructed or manufactured subjects. We extended ``artificial'' to include abstract images as well.

Secondary considerations subsequently shaped the image set subject to sampling the defined dimensions. We sought representation across a range of historical periods and geographic or cultural contexts. Public‑domain availability was a mandatory requirement. We made sure to include works containing distinct motifs or recognisable stylistic markers, as these features provide clearer signals for evaluating style transfer performance. Finally, we endeavoured to incorporate a range of difficulty levels for neural style transfer methods. Some images exhibit relatively straightforward painterly characteristics, such as textured brushwork without deliberate geometric distortion. At the same time, others present significant challenges, including rigid geometric structures, distinctive visual elements, or stylisations that deviate sharply from naturalistic representation. This diversity ensures that the evaluation can capture both the strengths and limitations of existing algorithms and will remain relevant as more capable methods emerge in the future. Discussion of individual style images appears in Appendix~\ref{sec:style image selection}.
\subsection{Content Image Selection}
\label{sec:content references selection}
We opted to draw from the already-curated
\textit{NPRgeneral}~\cite{mould2017developing} image set. These twenty images cover varied subject matter and visual elements, and collectively offer a wide range of challenges to
NPR methods. Unfortunately, it is impractical to use the full NPRgeneral set: the combinatorial explosion of style, content, and algorithm demands that we keep each category small. We selected six images, taking two images for each of the three content categories already established for style:
people, nature, and artificial.

Within the constraints implied by the above, we then chose a set of images that maximised coverage of the image properties described for the NPRgeneral images: colourful, high contrast, long gradients,
and many more. This process led us to the six images:
\textit{athletes} and \textit{mac} for people;
\textit{berries} and \textit{daisy} for nature;
\textit{angel} and \textit{barn} for artificial.

\subsection{Method Selection}
\label{sec:method selection}
There is a large literature on neural style transfer (we estimate 500–1000 published methods including all minor variations). To choose a manageable set of methods, we restricted the choice to papers with publicly available code, with a preference for influential papers, along with Multimodal Large Language Models for their generative capabilities different from NST methods.
We began with an initial pool of 20 candidate methods: AAMS~\cite{yao2019attention}, AdaAttN~\cite{liu2021adaattn}, AAST~\cite{hu2020aesthetic},AdaIN~\cite{huang2017arbitrary}, AesUST~\cite{wang2022aesust}, ArtFlow~\cite{an2021artflow}, CCPL~\cite{wu2022ccpl}, DiffuseST~\cite{hu2024diffusest}, Gatys~\cite{gatys2016image}, Gemini~\cite{team2023gemini}, ChatGPT~\cite{achiam2023gpt}, IEContraAST~\cite{chen2021artistic}, Linear~\cite{li2019learning}, OmniStyle~\cite{wang2025omnistyle}, RAST~\cite{ma2023rast}, SANET~\cite{park2019arbitrary}, StyTR-2~\cite{deng2022stytr2}, StyleID~\cite{chung2024style}, USO~\cite{wu2025uso}, NST-Ghiasi~\cite{ghiasi2017exploring}).
To this, we added a suite of handcrafted algorithms based on low-level image processing (LLIP)~\cite{jacobs2001image,tomasi1998bilateral,faraj2017generic,tschumperle2023g,rosin2017watercolour,blusseau2022adaptive,xu2011image,lu2012combining,berlin,kang2007coherent,pitie2007automated}, one per target style;
details appear in Appendix~\ref{sec:Details of LLIP Method Construction}.

To finalise the method selection from the candidate pool, we compute pairwise distances between methods using clean-FID~\cite{parmar2022aliased}.
For each style, clean-FID is computed between the outputs of two methods under the same content–style conditions, and the final distance is obtained by averaging across all styles.
For each method, we then compute the mean distance to all other methods as a coarse measure of its distinctiveness.
We were then able to exclude methods that did not offer enough variety, as measured by low pairwise distances. This step aims to reduce redundancy while retaining representative approaches. Following this process, we retain a final set of 10 NST methods: AAMS~\cite{yao2019attention}, AdaAttN~\cite{liu2021adaattn}, AdaIN~\cite{huang2017arbitrary}, ArtFlow~\cite{an2021artflow}, Gatys~\cite{gatys2016image}, OmniStyle~\cite{wang2025omnistyle}, SANET~\cite{park2019arbitrary}, StyTR-2~\cite{deng2022stytr2}, USO~\cite{wu2025uso}, NST-Ghiasi~\cite{ghiasi2017exploring},
plus ChatGPT~\cite{achiam2023gpt} and the handmade LLIP methods.
This process ensured a high level of diversity among the methods. Examples of selected images in ASTRA-Data produced by different algorithms are shown in Fig.~\ref{fig:examples_of_selection}.
\begin{table}[t]
    \centering
    \caption{Number of appearances of each method in Stage~1 and Stage~2 pairwise comparisons.}

    \label{tab:method_appearances_combined}
    \begin{tabular}{lccc}
        \toprule
        Method & Stage~1 & Stage~2 & Total \\
        \midrule
        AAMS~\cite{yao2019attention}       & 2886 & 2712 & 5598 \\
        AdaAttN~\cite{liu2021adaattn}     & 2983 & 2854 & 5837 \\
        AdaIN~\cite{huang2017arbitrary}   & 3026 & 2845 & 5871 \\
        ArtFlow~\cite{an2021artflow}     & 2960 & 2855 & 5815 \\
        ChatGPT~\cite{achiam2023gpt}     & 2836 & 2771 & 5607 \\
        Gatys~\cite{gatys2016image}       & 3052 & 2859 & 5911 \\LLIP%
        & 2869 & 2887 & 5756 \\
        OmniStyle~\cite{wang2025omnistyle} & 2933 & 2888 & 5821 \\
        SANET~\cite{park2019arbitrary}    & 2903 & 2815 & 5718 \\
        StyTR-2~\cite{deng2022stytr2}    & 3034 & 2741 & 5775 \\
        USO~\cite{wu2025uso}             & 2919 & 2883 & 5802 \\
        NST-Ghiasi~\cite{ghiasi2017exploring}  & 2735 & 2970 & 5705 \\
        \bottomrule
    \end{tabular}
\end{table}
\section{Annotation and Preference Analysis}
We conducted a two-stage online user study on 864 stylised images obtained from our data selection process, with the goal of constructing a human-centred preference dataset. These pairwise judgements were subsequently aggregated using Rank Centrality~\cite{negahban2017rank}, enabling us to recover a global ranking over all stylised images as well as corresponding preference scores for content preservation, style similarity, and overall quality. The user study was conducted under approval from the relevant institutional research ethics committee. Participants were recruited through open channels, resulting in 407 submissions. After applying quality control procedures, 264 valid responses were retained for analysis. Participants were compensated upon successful completion of the study.

This section presents the design and execution of the two-stage user study (Section~\ref{sec:survey interface}), followed by the procedures used to ensure data quality and annotation reliability (Section~\ref{sec:User Study Protocol}). We then examine the adequacy of the collected pairwise comparisons for stable global ranking through simulation (Section~\ref{sec:simulation}). Building on the validated annotations, we describe the construction of preference scores from raw pairwise data (Section~\ref{sec:groundtruth_establishment}) and subsequently analyse the resulting human preference patterns (Section~\ref{sec:data analysis}).

\subsection{Two-Stage Study Design and Interface}
\label{sec:survey interface}
In \textbf{Stage 1}, we perform controlled pairwise comparisons between different methods under the same content–style pair, isolating the effect of the generation method and yielding reliable within-pair preferences. Our design follows the \emph{two-alternative forced choice} (2AFC) protocol, which provides precise and consistent measurements of aesthetic preferences in style transfer~\cite{so2023measuring}.
Compared to absolute evaluations such as Likert-scale assessments of image quality, 2AFC admits finer gradations to be expressed, and eliminates problems arising from scale inconsistency among participants. The ``forced choice'' aspect ensures that participants provide a judgement for every pair, and cannot prematurely abandon the decision by selecting a deliberately excluded ``no preference'' option.
In Stage~1, we collected 17,568 votes from 122 participants.

Although each method appears in every content–style group, Stage~1 comparisons are local (within the same content-style pairs), preventing a  globally consistent ranking for method comparison and metric evaluation.
We collect data for global evaluations in \textbf{Stage 2}, where participants judge pairwise comparisons across different content–style pairs. These cross-category comparisons enable a coherent global ranking over all stylised images.
The Stage~2 comparisons are perceptually more challenging than those in Stage~1, so each questionnaire in Stage~2 contains fewer tasks. In Stage~2, we collected 17,040 votes from 142 participants, resulting in 34,608 total votes.
The user interface is illustrated in Figure~\ref{fig:userstudy_interface}.
\subsection{User Study Protocol and Quality Control}
\label{sec:User Study Protocol}
The user study first introduced the task and explained the style transfer concept, with examples and sample tasks.
Participants read an illustrated page explaining the key concepts of \emph{content}, \emph{style}, and \emph{style transfer} using text and visual examples. They then completed a short interactive tutorial with representative examples to familiarise themselves with the evaluation criteria and question format (see details in Appendix~\ref{sec:user_study_training}). On completing the tutorial, the participant begins answering questions about style transfer results from different algorithms. Their answers constitute the user preference data which we record and later analyse.

In both Stage~1 and Stage~2 studies, each comparison required participants to answer three questions:
(1) which stylisation result looks more like an ideal combination of the content and the style?
(2) which result better captures the style? and
(3) which result better preserves the content?

In Stage~1, each comparison is obtained by selecting a random style image, a random content image, and two random algorithms;
the reference content and style images are shown to the user together with the
two stylised results. Thus each comparison comprises three questions based on a fixed content-style pair. Multiple comparisons within a questionnaire are randomly sampled without repetition.
This setup yields reliable \emph{within-group} preferences under controlled conditions. Stage~2 consisted of cross-group comparisons, connecting different content-style pairs.
Each question contained two randomly sampled stylised results together with their respective content and style references, yielding six images in total, i.e., $(c_1, s_1, m_1)$ and $(c_2, s_2, m_2)$. The user was asked to evaluate which of $m_1$ or $m_2$ better reflected the criteria. For content, is $m_1$ closer to $c_1$ than $m_2$ is to $c_2$? For
style, is $m_1$ closer to $s_1$ than
$m_2$ is to $s_2$? For overall preference, does $m_1$ better reflect the $(c_1,s_1)$ style transfer than $m_2$ does for $(c_2,s_2)$?

\noindent {\color{black}
\textbf{Annotation consistency.}
To further assess annotation reliability, we re-annotated a randomly selected subset of comparisons using additional independent annotators. In Stage~1, Cohen's $\kappa$ values were 0.663, 0.547, and 0.547 for content, style, and overall preference, respectively. In Stage~2, the corresponding values were 0.515, 0.389, and 0.416. These results show meaningful but task-dependent agreement, with stronger consistency in Stage~1 and reduced agreement in the more demanding cross-pair comparisons of Stage~2. Further details are provided in Appendix~\ref{app:annotation_consistency}.
}
\subsection{Two-Stage Ranking Stability Simulation}
\label{sec:simulation}
We analyse whether the number of pairwise comparisons in our two-stage study was sufficient to support stable ranking estimation.
Spectral ranking theory~\cite{negahban2017rank,chen2015spectral} suggests that reliable recovery scales as $\Theta(n\log n)$ under models such as Bradley-Terry-Luce~\cite{bradley1952rank,luce1959individual}, providing a theoretical reference for the required comparison scale.
We further validate this empirically via simulation using Rank Centrality~\cite{negahban2017rank}.
For Stage~1, ranking stability rapidly converges and exceeds $\rho=0.99$, where $\rho$ denotes the Spearman rank correlation between the estimated ranking and the reference ranking, at approximately $1.2\times10^4$ comparisons.
For Stage~2, global ranking accuracy improves steadily with additional cross-group comparisons, reaching $\rho\approx0.95$ at around $1.4\times10^4$ comparisons.
In practice, our study collected 17,568 (Stage~1) and 17,040 (Stage~2) votes, both of which lie well within the stable regime. See more details in Appendix~\ref{sec:simulation_appendix}.

\noindent \textbf{Empirical graph diagnostics.}
We additionally examine the topology of the comparison graph constructed from the actual collected annotations. As expected, Stage~1 alone forms 72 disconnected components corresponding to the 12 style references and 6 content references. The cross-group comparisons introduced in Stage~2 connect these components, yielding a single connected comparison graph over all 864 stylised results. The combined graph has an average node
degree of approximately 45.4. The Rank Centrality transition matrices for content, style, and overall preference have spectral gaps of 0.0199, 0.0421, and 0.1000, respectively. We further perform 100 stratified bootstrap resamplings of the collected comparisons; all resampled graphs remain connected and Rank Centrality can be successfully computed in
every case. Together with the ranking-recovery simulation above, these diagnostics confirm that the global rankings are not obtained from disconnected comparison components and remain stable under resampling. Further connectivity and bootstrap analyses are provided in Appendix~\ref{sec:graph_diagnostics}.
\subsection{Preference Score Estimation}
\label{sec:groundtruth_establishment}
To obtain reliable preference scores, we combine pairwise comparisons from both stages of the user study.
Stage~1 provides comparisons across methods under identical $(\text{content}, \text{style})$ conditions, while Stage~2 connects different content–style pairs to enable meaningful global rankings.

Participants provide pairwise preferences under three criteria (overall, style, content), defining three separate ranking problems. Our goal is to recover a global preference ranking for each criterion. Each generated image is treated as a player:
\begin{equation}
p = (\text{style}, \text{content}, \text{method}).
\end{equation}

\noindent \textbf{Ranking from Pairwise Comparisons.}
Given these pairwise comparisons between players, we adopt \textbf{Rank Centrality}~\cite{negahban2017rank} to estimate global rankings.
Rank Centrality is a spectral method that interprets comparison outcomes as transitions of a Markov chain and recovers rankings from its stationary distribution.

\noindent \textbf{Rank Centrality Formulation.}
Let $w_{ij}$ denote the number of times player $i$ defeats player $j$, and
\begin{equation}
n_{ij} = w_{ij} + w_{ji}
\end{equation}
be the total number of comparisons between the pair.
Following~\cite{negahban2017rank}, we construct a transition matrix $P$ for a Markov chain where the transition probability from player $i$ to player $j$ is

\begin{equation}
P_{ij} =
\begin{cases}
\frac{1}{d_{\max}} \frac{w_{ji}}{n_{ij}}, & n_{ij} > 0, \ i \neq j,\\
1 - \sum_{k \neq i} P_{ik}, & i = j,
\end{cases}
\end{equation}
where $d_{\max}$ denotes the maximum degree of the comparison graph.
Intuitively, the transition probability is proportional to the empirical probability that player $j$ defeats player $i$.
The stationary distribution $\boldsymbol{\pi}$ of this Markov chain satisfies

\begin{equation}
\boldsymbol{\pi} = \boldsymbol{\pi} P ,
\end{equation}
and the resulting stationary probability $\pi_i$ is taken as the preference score of player $i$.
Players with larger stationary probabilities correspond to having more preferred stylisation results.

\begin{figure}[h]
    \centering
    \includegraphics[width=\linewidth]{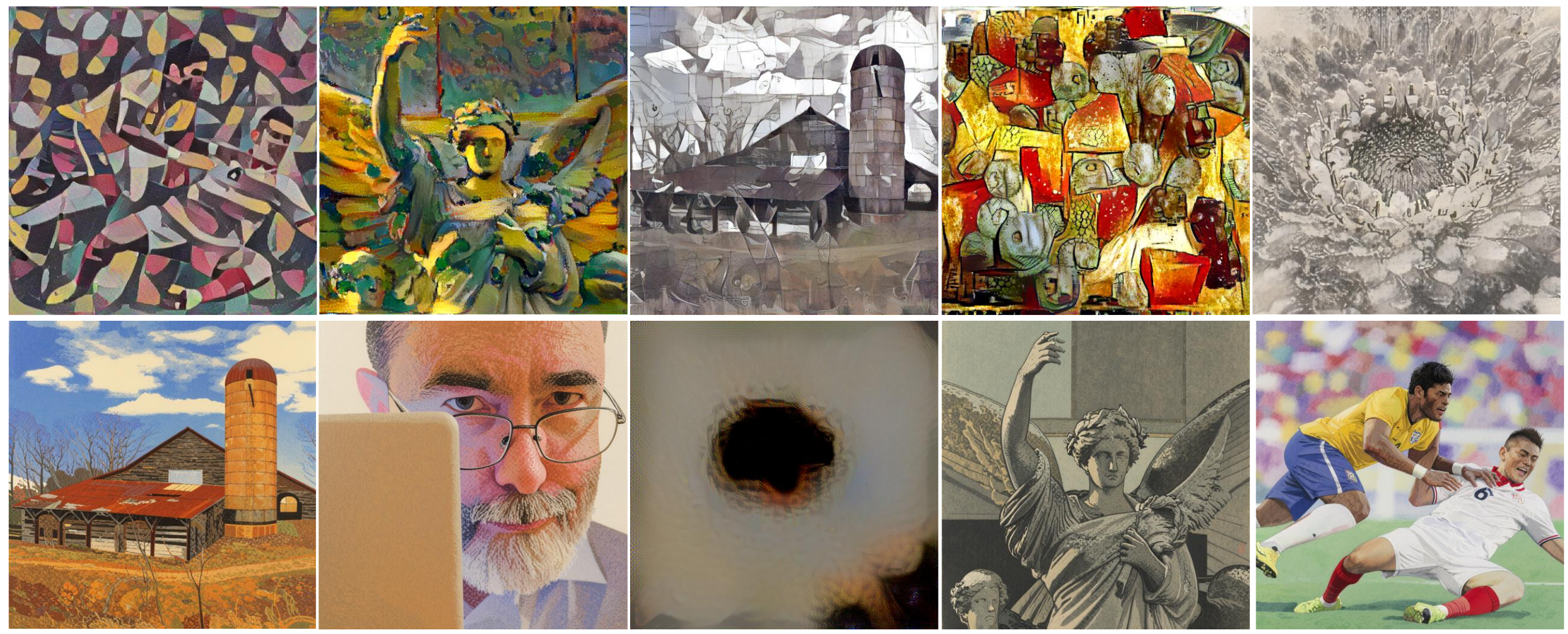}
    \caption{
    Top-5 (top row) and bottom-5 (bottom row) stylised results selected from the top-10 and bottom-10 samples
    ranked by their mean Rank Centrality (RC) scores for overall preference.
    The style references for each column, arranged from left to right and top to bottom, are:
    Kandinsky, Carr, Picasso, Ptolemaic Egyptian, Chinese landscape, Australian Aboriginal, Australian Aboriginal, Rembrandt, Hokusai, and Picasso.
    }

    \label{fig:Overall_top5_bottom5}
\end{figure}
\subsection{Preliminary Data Analysis}
\label{sec:data analysis}
\textbf{Raw Pairwise Data.}
Based on the collected data, we compute the frequency with which each stylisation method appears in the pairwise comparison tasks.
After filtering incomplete responses and failed attention checks, we aggregate the occurrences of each method across all valid comparisons.
Tab.~\ref{tab:method_appearances_combined} summarises the number of appearances in Stage~1 and Stage~2.
Overall, the exposure of different methods is approximately balanced.
In Stage~1, each method appears about $2.7\mathrm{k}$–$3.1\mathrm{k}$ times, while in Stage~2 the counts range from $2.6\mathrm{k}$ to $3.0\mathrm{k}$.
Similarly, each style reference participates in roughly $200$–$320$ comparisons per method, and each content image appears about $400$–$560$ times per method (see more details in Appendix~\ref{app:exposure}).
This balanced distribution ensures that the pairwise comparison data are well distributed across methods, styles, and content, preventing the final preference estimation from being dominated by a small subset of references.

\begin{figure}[t]
    \centering
    \begin{minipage}[t]{0.49\linewidth}
        \centering
        \includegraphics[width=\linewidth]{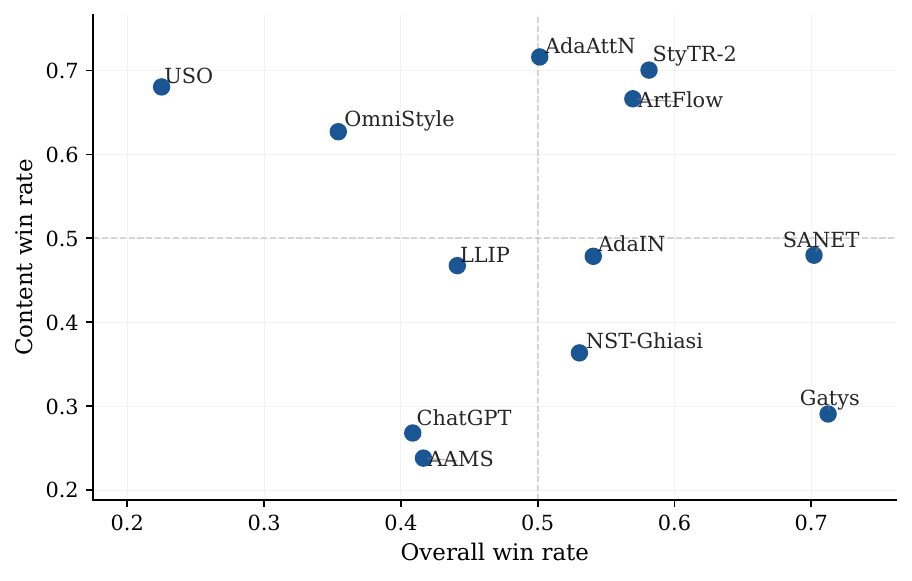}
        \caption*{(a) Overall vs Content}
    \end{minipage}
    \hfill
    \begin{minipage}[t]{0.49\linewidth}
        \centering
        \includegraphics[width=\linewidth]{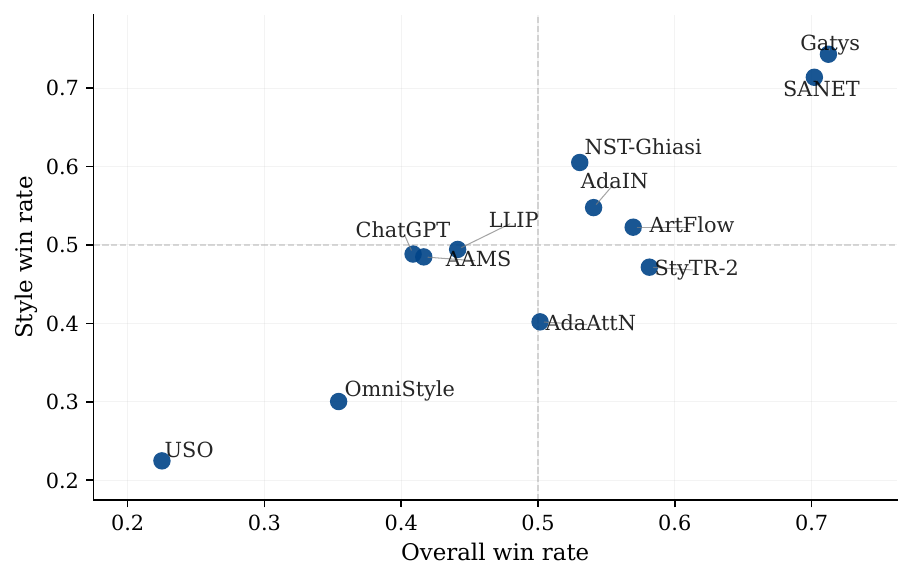}
        \caption*{(b) Overall vs Style}
    \end{minipage}
    \caption{Scatter plots of (a) overall effectiveness vs. content preservation; (b) overall effectiveness vs. style fidelity.}

    \label{fig:twoscatter}
\end{figure}

\begin{figure}[t]
    \centering
    \includegraphics[width=\linewidth]{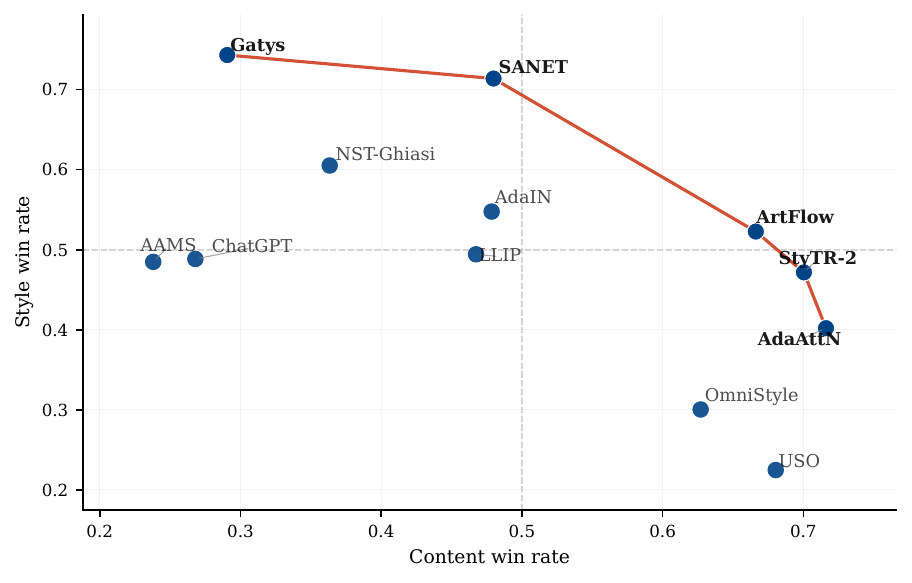}
    \caption{
    Trade-off between style similarity and content preservation measured from Stage~1 pairwise comparisons.
    The dashed curve indicates the Pareto frontier of the two objectives, which highlights the optimal trade-off boundary between the two criteria.
    }

    \label{fig:method_tradeoff}
\end{figure}
\label{sec:method_tradeoff}

For
each method $m$, we compute preference rates from the votes: the \emph{style win rate}, \emph{content win rate},
and \emph{overall win rate},
defined as the proportion of comparisons in which the stylised result generated by $m$ is chosen in the corresponding question.
Fig.~\ref{fig:twoscatter} and
Fig.~\ref{fig:method_tradeoff}
show
scatter plots of combinations of these quantities for all methods: overall vs. content, overall vs. style, and
content vs. style.

Fig.~\ref{fig:twoscatter} (a) shows little pattern: participants' assessment of the overall effectiveness of style transfer was not strongly influenced by content preservation. In contrast,
Fig.~\ref{fig:twoscatter} (b) shows a strong positive correlation between
overall effectiveness of style transfer and style fidelity. Together, these two scatter plots indicate that
style fidelity is the most critical aspect of the style transfer task, with content preservation less clearly contributing.

Fig.~\ref{fig:method_tradeoff} helps us assess tradeoffs between style similarity and content preservation.
An idealised method will exhibit no tradeoff: both content and style can be preserved. However, in practice, tradeoffs do exist, with some methods favouring style, some methods favouring content. Furthermore, the tradeoffs are style-specific: some restrictive styles, such as Mondrian, are not conducive to presenting photographic content, while painterly styles aiming at greater realism will allow both style and content to be captured in the output.

In practice, there is substantial tension between the two objectives: methods achieving stronger style similarity (e.g., Gatys~\cite{gatys2016image}, SANET~\cite{park2019arbitrary}) tend to exhibit lower content fidelity, while methods emphasising content preservation (e.g., AdaAttN~\cite{liu2021adaattn}, StyTR-2~\cite{deng2022stytr2}) typically produce weaker stylistic effects.

\begin{figure}[ht]
    \centering

    \begin{subfigure}[t]{0.48\linewidth}
        \centering
        \includegraphics[width=\linewidth]{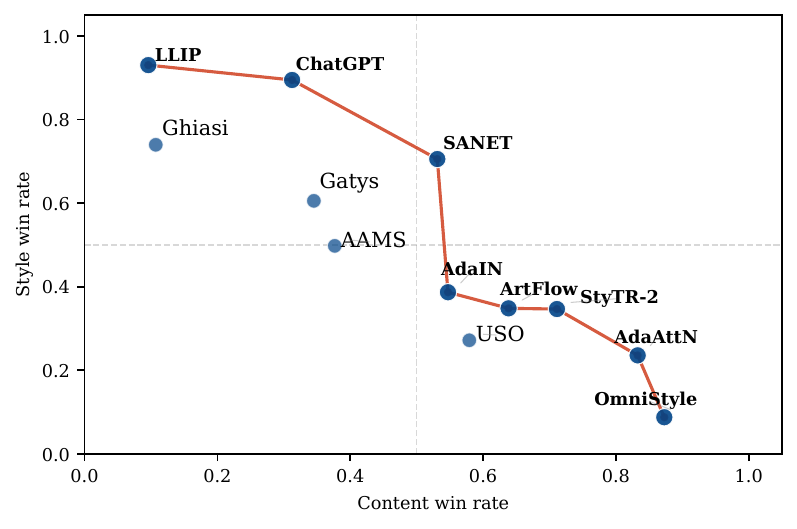}
        \caption{Mondrian Style}

    \end{subfigure}
    \hfill
    \begin{subfigure}[t]{0.48\linewidth}
        \centering
        \includegraphics[width=\linewidth]{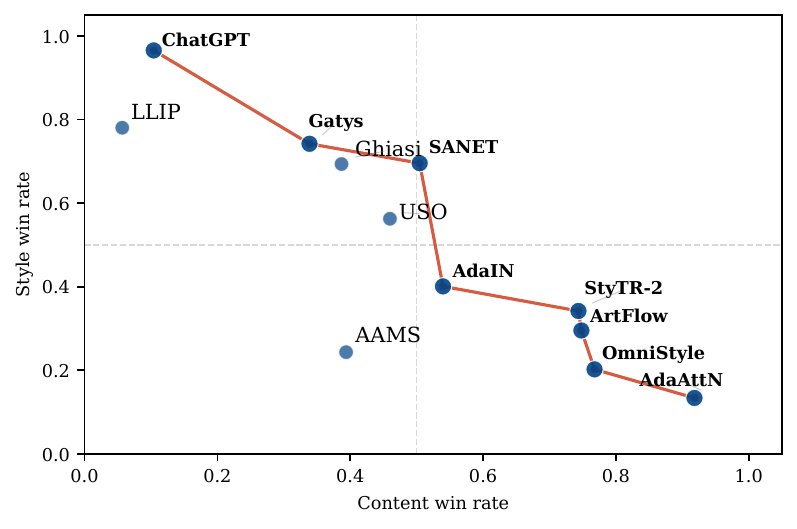}
        \caption{Kandinsky Style}

    \end{subfigure}
    \begin{subfigure}[t]{0.48\linewidth}
        \centering
        \includegraphics[width=\linewidth]{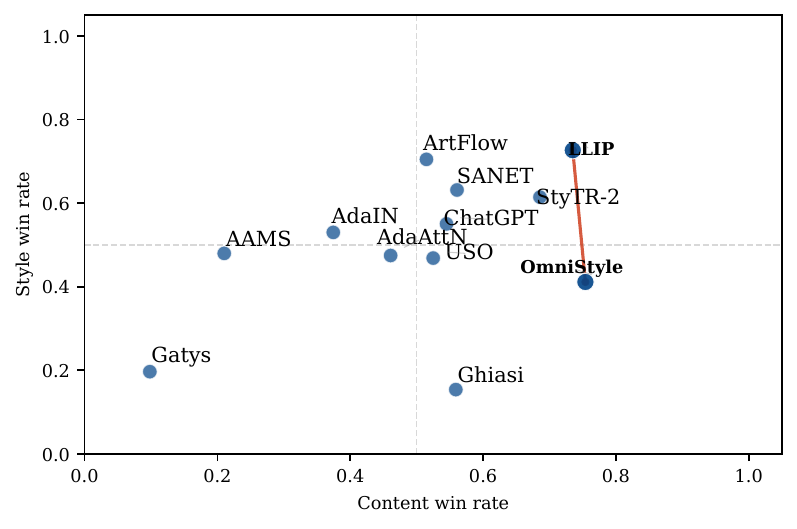}
        \caption{Rembrandt Style}

    \end{subfigure}
    \hfill
    \begin{subfigure}[t]{0.48\linewidth}
        \centering
        \includegraphics[width=\linewidth]{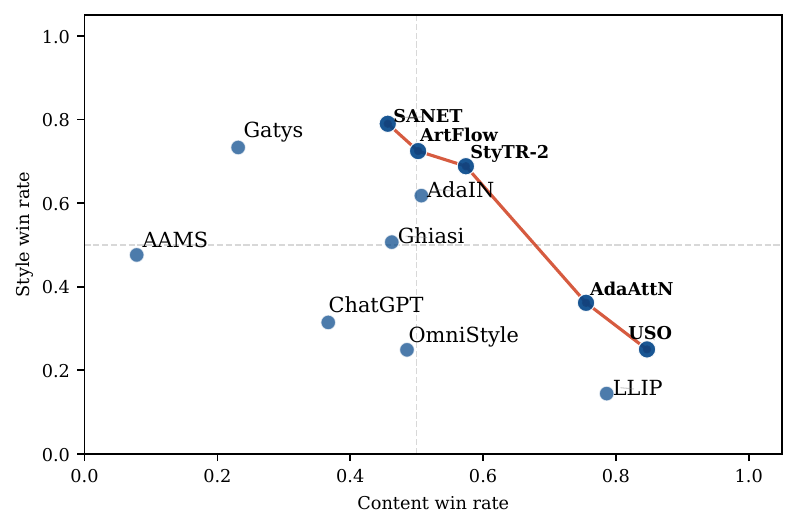}
        \caption{Bierstadt Style}

    \end{subfigure}
    \caption{
    Content–style preference trade-off under different style references.
    }
    \label{fig:style_tradeoff}
\end{figure}
To further examine whether this trade-off depends on style or content, we compute the same statistics per style and per content, and visualise the resulting distributions (see Appendix~\ref{sec:pareto_curve}).
The overall trend remains largely consistent across different content images.
In contrast, the behaviour varies substantially across styles.
Abstract styles (Figs.~\ref{fig:style_tradeoff} (a,b)) possess a clear inverse relationship between style and content preferences, where methods achieving strong stylistic effects often incur significant structural distortion and hence have reduced content preservation.

Not all styles demand such tradeoffs. Indeed, in Fig.~\ref{fig:style_tradeoff} (c) we can see a positive correlation between content and style preferences, indicating that better-performing algorithms can preserve both style and content effectively. Conversely, Fig.~\ref{fig:style_tradeoff} (d) lacks any clear trend. The degree to which tradeoffs are present remains style-dependent.

\begin{figure*}[t]
    \centering
    \begin{subfigure}[t]{0.32\textwidth}
        \centering
        \includegraphics[width=\linewidth]{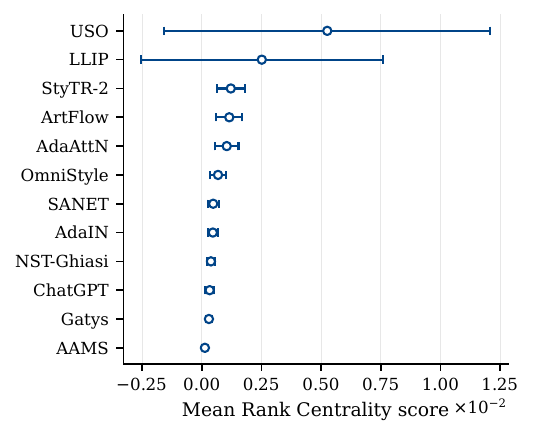}
        \caption{Content preference}

    \end{subfigure}
    \hfill
    \begin{subfigure}[t]{0.32\textwidth}
        \centering
        \includegraphics[width=\linewidth]{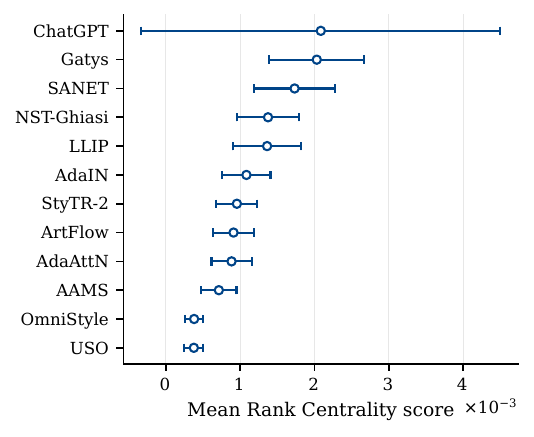}
        \caption{Style preference}

    \end{subfigure}
    \hfill
    \begin{subfigure}[t]{0.32\textwidth}
        \centering
        \includegraphics[width=\linewidth]{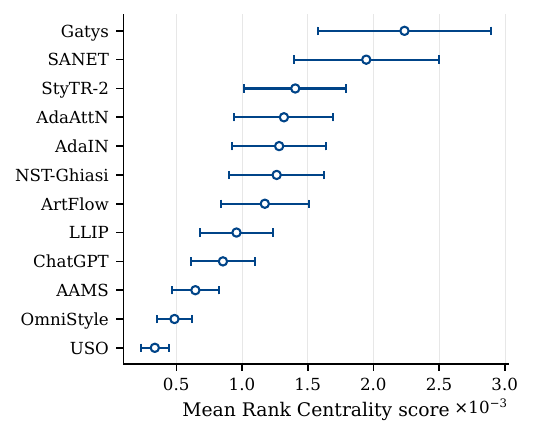}
        \caption{Overall preference}

    \end{subfigure}
    \caption{
    Human preference rankings derived from Rank Centrality.
    Each point shows the mean score and the error bar denotes bootstrap uncertainty.
    ``Overall preference'' refers to the subjects' holistic assessment of the effectiveness of the style transfer.
    }

    \label{fig:ranking_results}
\end{figure*}

\noindent \textbf{Preference Scores.} As illustrated in Fig.~\ref{fig:ranking_results}, Gatys~\cite{gatys2016image} achieves the highest Mean Rank Centrality score in overall preference, though it exhibits significant variance across different style-content pairs, as indicated by its wide error bar. In contrast, while newer methods like AdaAttN~\cite{liu2021adaattn} and AdaIN~\cite{huang2017arbitrary} show competitive and more stable performance, their overall stylisation effectiveness still falls short. Methods such as USO~\cite{wu2025uso} and OmniStyle~\cite{wang2025omnistyle} demonstrate high consistency (narrow error bars) but fail to achieve competitive scores in our evaluation framework. To further illustrate the qualitative variability of each method, we provide the top-3 and bottom-3 overall preference examples for every method in the Appendix~Fig.~\ref{fig:top3_bottom3_each_method}.

Importantly, we observe that overall preference aligns more closely with style similarity than with content preservation.
For instance, USO~\cite{wu2025uso} achieves the highest ranking in content, yet ranks among the lowest in both overall and style evaluations, revealing a clear discrepancy between content fidelity and success at style transfer.

To further understand this phenomenon, we analyse cases where content and style preferences disagree. Across a total of $18{,}747$ such conflict cases, participants follow the style preference in $80.3\%$ of comparisons, compared to $19.7\%$ for content. This trend aligns with the observations from Fig.~\ref{fig:twoscatter}, where the overall preference ranking closely aligns with style preference across methods.
The evidence supports the hypothesis that for the style transfer task, humans are more sensitive to discrepancies in the style than in the content.

Fig.~\ref{fig:Overall_top5_bottom5} shows the top-5 and bottom-5 results ranked by overall preference scores.
Top-ranked results often correspond to styles with strong and distinctive visual cues (e.g., Chinese landscape, Carr), suggesting that clear colour and texture patterns are more readily perceived and evaluated.
In contrast, bottom-ranked results exhibit two common failure modes: severe content degradation, where structure becomes unrecognisable (e.g., Rembrandt), and weak stylisation, where content is preserved but stylistic cues are minimal (e.g., Australian Aboriginal, Hokusai).
These observations indicate that both recognisable content and perceptually salient style cues are important, with the latter playing a particularly prominent role.

\section{ASTRA-Score and Experiments}
Next, we introduce \textbf{ASTRA-Score}, an evaluation model trained on \textbf{ASTRA-Data}, and compare it with existing metrics on the benchmark image set.
\subsection{Evaluation Model}
\label{sec:evaluation_model}
\begin{figure}[t]
    \centering
    \includegraphics[width=\linewidth]{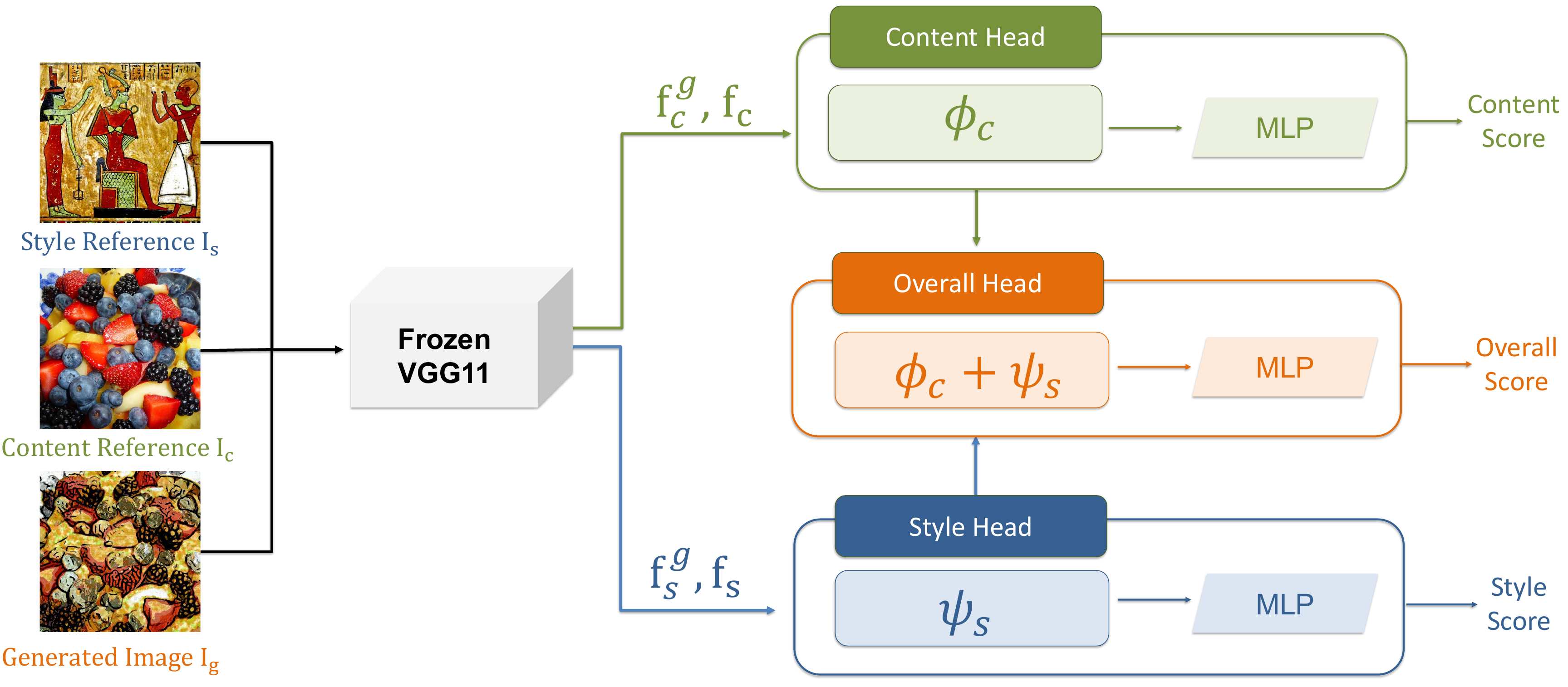}
    \caption{Overview of the proposed evaluation model.
    A frozen VGG11 extracts content features ($\mathbf{f}_c^g, \mathbf{f}_c$) and style features ($\mathbf{f}_s^g, \mathbf{f}_s$) from the generated image $I_g$, content reference $I_c$, and style reference $I_s$.
    Three independent heads are employed: the Content Head uses $\phi(\mathbf{f}_c^g,\mathbf{f}_c)$, the Style Head uses $\psi(\mathbf{f}_s^g,\mathbf{f}_s)$, and the Overall Head combines both relation features to predict an overall quality score.
    }

    \label{fig:architecture}
\end{figure}

To enable automatic assessment of stylised images, we learn a regression-based evaluation model that predicts human-aligned scores from a content-style-stylisation image triplet, as illustrated in Fig.~\ref{fig:architecture}.
Given a content reference image $I_c$, a style reference image $I_s$, and a generated stylised image $I_g$, the goal is to estimate how well the generated result preserves the content structure and matches the target style, and forms an ideal combination of the two.
We train three separate predictors corresponding to three evaluation criteria: \emph{overall preference}, \emph{content preservation}, and \emph{style similarity}.
Using the fully trained model, we can evaluate new stylisation methods on the benchmark image set.
We employ a frozen pretrained backbone to extract hierarchical visual
representations and train only lightweight task-specific regression heads.
We empirically compare several backbone families, including MobileNetV3,
VGG, ResNet, ViT, and DINOv2. Among the evaluated architectures, VGG11
provides the strongest overall alignment with human judgements across the
three evaluation criteria, and is therefore adopted as the default backbone
for ASTRA-Score. The backbone remains frozen throughout training, limiting
the number of trainable parameters to the regression heads. Full results of the backbone comparison are provided in Appendix~\ref{app:backbone_ablation}.

\noindent\textbf{Feature Representation. }
Given an input image $I$, we extract intermediate feature maps from four
progressively deeper stages of VGG11, denoted as
$\{L_i\}_{i=1}^{4}$, with channel dimensions of 64, 128, 256, and 512,
respectively.

\noindent\textbf{Content and Style Features.}
The deepest feature map $L_4$ is used to represent semantic content.
We apply Global Average Pooling (GAP) to obtain a content feature vector
$\mathbf{f}_c \in \mathbb{R}^{512}$.
For style representation, we compute the channel-wise mean
$\mu(\cdot)$ and standard deviation $\sigma(\cdot)$ for each selected
feature map $L_i$~\cite{liu2021adaattn}.
The style descriptor at each level is defined as
$v_i=[\mu(L_i)^\top,\sigma(L_i)^\top]^\top$.
The final multi-level style representation is obtained by concatenating
the statistics from all four stages,
\begin{equation}
\mathbf{f}_s =
[v_1^\top,v_2^\top,v_3^\top,v_4^\top]^\top
\in \mathbb{R}^{1920}.
\end{equation}\mbox{}

\noindent\textbf{Relation features.}
We model the relationship between the generated image and the references in feature space.
Let $\mathbf{f}_c^g,\mathbf{f}_c$ denote content features, and $\mathbf{f}_s^g,\mathbf{f}_s$ denote style features.
We define two relation operators:
\begin{equation}
\phi(\mathbf{a},\mathbf{b})=
\left[
|\mathbf{a}-\mathbf{b}|,\;
\mathbf{a}\odot\mathbf{b},\;
\mathrm{cos}(\mathbf{a},\mathbf{b})
\right],
\end{equation}
\begin{equation}
\psi(\mathbf{a},\mathbf{b})=
\left[
|\mathbf{a}-\mathbf{b}|,\;
\mathbf{a}\odot\mathbf{b},\;
\mu(|\mathbf{a}-\mathbf{b}|),\;
\mu((\mathbf{a}-\mathbf{b})^2)
\right],
\end{equation}
where $\mu(\cdot)$ denotes the mean over feature dimensions.
The three predictors use task-specific relation features:
\begin{equation}
\mathbf{x}_{\mathrm{content}}=\phi(\mathbf{f}_c^g,\mathbf{f}_c), \quad
\mathbf{x}_{\mathrm{style}}=\psi(\mathbf{f}_s^g,\mathbf{f}_s),
\end{equation}
\begin{equation}
\mathbf{x}_{\mathrm{overall}}=
\left[
\phi(\mathbf{f}_c^g,\mathbf{f}_c),\;
\psi(\mathbf{f}_s^g,\mathbf{f}_s)
\right].
\end{equation}
\textbf{Regression heads.}
Each predictor is implemented as a lightweight two-layer MLP with dropout, producing a scalar score.
The models are trained independently using
3 preference scores for each evaluation dimension, with full training details provided in Appendix~\ref{sec:hyperparameters}.

\subsection{Evaluation Methods Comparison}
\begin{table}[t]
    \centering
    \caption{Correlation between automatic metrics and study-derived rankings across the evaluated methods.
    Baseline metrics are evaluated using leave-one-method-out (LOMO) evaluation and reported as mean $\pm$ standard deviation.}
    \label{tab:metric_correlation}

    \begin{tabular}{lcc}
        \toprule
        Metric & Spearman & Kendall \\
        \midrule

        \rowcolor{gray!10}
        \multicolumn{3}{l}{\textbf{Content}} \\

        CFSD~\cite{chung2024style} & 0.281 $\pm$ 0.212 & 0.196 $\pm$ 0.151 \\
        NMI~\cite{studholme1998normalized} & 0.415 $\pm$ 0.141 & 0.288 $\pm$ 0.100 \\
        SIFID~\cite{shaham2019singan} & 0.429 $\pm$ 0.170 & 0.300 $\pm$ 0.129 \\
        GMSD~\cite{xue2013gradient} & 0.456 $\pm$ 0.176 & 0.323 $\pm$ 0.133 \\
        FSIM~\cite{zhang2011fsim} & 0.515 $\pm$ 0.160 & 0.368 $\pm$ 0.118 \\
        Qwen~\cite{bai2023qwen} & 0.526 $\pm$ 0.212 & 0.414 $\pm$ 0.168 \\
        VIF~\cite{sheikh2006image} & 0.551 $\pm$ 0.178 & 0.396 $\pm$ 0.136 \\
        CLIP~\cite{radford2021learning} & 0.559 $\pm$ 0.165 & 0.401 $\pm$ 0.138 \\
        MS-SSIM~\cite{wang2003multiscale} & 0.582 $\pm$ 0.137 & 0.416 $\pm$ 0.104 \\
        LPIPS~\cite{zhang2018unreasonable} & 0.586 $\pm$ 0.129 & 0.422 $\pm$ 0.108 \\
        DINO~\cite{caron2021emerging} & 0.610 $\pm$ 0.160 & 0.442 $\pm$ 0.132 \\
        ViT~\cite{dosovitskiy2020image} & 0.614 $\pm$ 0.155 & 0.444 $\pm$ 0.130 \\
        SRQE~\cite{chen2022quality} & 0.634 $\pm$ 0.150 & 0.461 $\pm$ 0.119 \\
        \textbf{ASTRA-Score Content (ours)} & \textbf{0.718 $\pm$ 0.144} & \textbf{0.540 $\pm$ 0.128} \\

        \midrule
        \rowcolor{gray!10}
        \multicolumn{3}{l}{\textbf{Style}} \\

        Gram Loss~\cite{gatys2016image} & 0.039 $\pm$ 0.240 & 0.030 $\pm$ 0.166 \\
        CSD~\cite{somepalli2024measuring} & 0.066 $\pm$ 0.268 & 0.047 $\pm$ 0.185 \\
        SIFID~\cite{shaham2019singan} & 0.103 $\pm$ 0.128 & 0.073 $\pm$ 0.090 \\
        DINO~\cite{caron2021emerging} & 0.152 $\pm$ 0.281 & 0.111 $\pm$ 0.190 \\
        CLIP~\cite{radford2021learning} & 0.199 $\pm$ 0.162 & 0.137 $\pm$ 0.109 \\
        Qwen~\cite{bai2023qwen} & 0.250 $\pm$ 0.204 & 0.190 $\pm$ 0.153 \\
        ViT~\cite{dosovitskiy2020image} & 0.290 $\pm$ 0.137 & 0.198 $\pm$ 0.098 \\
        SRQE~\cite{chen2022quality} & 0.390 $\pm$ 0.207 & 0.274 $\pm$ 0.149 \\
        \textbf{ASTRA-Score Style (ours)} & \textbf{0.714 $\pm$ 0.098} & \textbf{0.531 $\pm$ 0.084} \\

        \midrule
        \rowcolor{gray!10}
        \multicolumn{3}{l}{\textbf{Overall}} \\

        ArtFID~\cite{wright2022artfid} & 0.109 $\pm$ 0.173 & 0.076 $\pm$ 0.117 \\
        Qwen~\cite{bai2023qwen} & 0.295 $\pm$ 0.172 & 0.233 $\pm$ 0.136 \\
        SRQE~\cite{chen2022quality} & 0.356 $\pm$ 0.223 & 0.247 $\pm$ 0.154 \\
        \textbf{ASTRA-Score Overall (ours)} & \textbf{0.667 $\pm$ 0.099} & \textbf{0.492 $\pm$ 0.081} \\

        \bottomrule
    \end{tabular}
\end{table}

\begin{table}[ht]
    \centering
    \caption{Correlation between automatic metrics and study-derived rankings across the evaluated methods.
    Baseline metrics are evaluated using leave-one-content-out (LOCO) evaluation and reported as mean $\pm$ standard deviation.}
    \begin{tabular}{lcc}
        \toprule
        Metric & Spearman & Kendall \\
        \midrule

        \rowcolor{gray!10}
        \multicolumn{3}{l}{\textbf{Content}} \\
        CFSD~\cite{chung2024style}
            & 0.440 $\pm$ 0.055
            & 0.310 $\pm$ 0.044 \\
        Qwen~\cite{bai2023qwen}
            & 0.455 $\pm$ 0.184
            & 0.357 $\pm$ 0.145 \\
        NMI~\cite{studholme1998normalized}
            & 0.492 $\pm$ 0.043
            & 0.345 $\pm$ 0.033 \\
        GMSD~\cite{xue2013gradient}
            & 0.498 $\pm$ 0.103
            & 0.361 $\pm$ 0.081 \\
        FSIM~\cite{zhang2011fsim}
            & 0.544 $\pm$ 0.086
            & 0.395 $\pm$ 0.074 \\
        SIFID~\cite{shaham2019singan}
            & 0.557 $\pm$ 0.137
            & 0.400 $\pm$ 0.102 \\
        VIF~\cite{sheikh2006image}
            & 0.558 $\pm$ 0.113
            & 0.417 $\pm$ 0.077 \\
        MS-SSIM~\cite{wang2003multiscale}
            & 0.577 $\pm$ 0.043
            & 0.419 $\pm$ 0.024 \\
        SRQE~\cite{chen2022quality}
            & 0.585 $\pm$ 0.075
            & 0.428 $\pm$ 0.049 \\
        LPIPS~\cite{zhang2018unreasonable}
            & 0.705 $\pm$ 0.079
            & 0.525 $\pm$ 0.070 \\
        ViT~\cite{dosovitskiy2020image}
            & 0.718 $\pm$ 0.087
            & 0.537 $\pm$ 0.080 \\
        CLIP~\cite{radford2021learning}
            & 0.721 $\pm$ 0.033
            & 0.528 $\pm$ 0.033 \\
        \textbf{DINO~\cite{caron2021emerging}}
            & \textbf{0.759 $\pm$ 0.082}
            & \textbf{0.573 $\pm$ 0.071} \\
        ASTRA-Score Content (ours) & 0.745 $\pm$ 0.094 & 0.562 $\pm$ 0.090 \\

        \midrule
        \rowcolor{gray!10}
        \multicolumn{3}{l}{\textbf{Style}} \\
        CSD~\cite{somepalli2024measuring}
            & 0.130 $\pm$ 0.094
            & 0.088 $\pm$ 0.063 \\
        CLIP~\cite{radford2021learning}
            & 0.139 $\pm$ 0.102
            & 0.092 $\pm$ 0.072 \\
        SIFID~\cite{shaham2019singan}
            & 0.190 $\pm$ 0.098
            & 0.126 $\pm$ 0.065 \\
        Gram Loss~\cite{gatys2016image}
            & 0.227 $\pm$ 0.084
            & 0.162 $\pm$ 0.056 \\
        Qwen~\cite{bai2023qwen}
            & 0.262 $\pm$ 0.055
            & 0.198 $\pm$ 0.047 \\
        DINO~\cite{caron2021emerging}
            & 0.378 $\pm$ 0.106
            & 0.261 $\pm$ 0.073 \\
        ViT~\cite{dosovitskiy2020image}
            & 0.472 $\pm$ 0.038
            & 0.328 $\pm$ 0.028 \\
        SRQE~\cite{chen2022quality}
            & 0.529 $\pm$ 0.023
            & 0.380 $\pm$ 0.017 \\
       \textbf{ASTRA-Score Style (ours)} & \textbf{0.814 $\pm$ 0.037} & \textbf{0.626 $\pm$ 0.036} \\

        \midrule
        \rowcolor{gray!10}
        \multicolumn{3}{l}{\textbf{Overall}} \\
        ArtFID~\cite{wright2022artfid}
            & 0.115 $\pm$ 0.115
            & 0.075 $\pm$ 0.072 \\
        Qwen~\cite{bai2023qwen}
            & 0.179 $\pm$ 0.105
            & 0.138 $\pm$ 0.081 \\
        SRQE~\cite{chen2022quality}
            & 0.545 $\pm$ 0.036
            & 0.390 $\pm$ 0.023 \\
        \textbf{ASTRA-Score Overall (ours)} & \textbf{0.793 $\pm$ 0.076} & \textbf{0.606 $\pm$ 0.080} \\
        \bottomrule
    \end{tabular}
    \label{tab:metric_correlation_contentwise}
\end{table}
We evaluate whether objective metrics can predict perceptual rankings on ASTRA-Data, which contains $12$ styles, $6$ contents, and $12$ stylisation methods.
To assess generalisation across methods, we adopt a leave-one-method-out protocol.
In each run, one method is held out while the remaining $11$ methods are used to train the prediction model, which then predicts the ranking of the held-out method over its $72$ generated images.
The predicted ranking is compared with the global rankings using Spearman’s $\rho$~\cite{spearman1961proof} and Kendall’s $\tau$~\cite{kendall1938new}.
This process is repeated for all $12$ methods, and the final performance is reported as the average correlation across runs.

We further evaluate a range of commonly used style transfer metrics, including style similarity, content preservation, and overall quality measures, under the same protocol.
For each metric, rankings are computed for the held-out method and compared with the ground truth in the same manner.
Tab.~\ref{tab:metric_correlation} reveals a clear gap between existing automatic metrics and perceptual preference. SRQE~\cite{chen2022quality} achieves relatively strong performance, demonstrating the value of human-annotated supervision among prior approaches. However, its reliance on fixed content–style pairs entangles scores with pair-specific difficulty, and its evaluator is still based on handcrafted features rather than directly learned from user evaluations.

Even the best prior automatic metrics achieve only moderate correlation with human rankings, while several widely used measures (e.g., CFSD~\cite{chung2024style} and SIFID (Single Image FID)~\cite{shaham2019singan}) exhibit weak correlation in the content evaluation, indicating that minimising perceptual distance does not necessarily lead to more preferred stylisations.

The discrepancy is even more pronounced for style and overall evaluations: classical statistics such as Gram Loss~\cite{gatys2016image} and recent pretrained vision-language model based metrics, including embedding-based similarity (e.g., CLIP~\cite{radford2021learning})
and instruction following models such as Qwen~\cite{bai2023qwen}, show low and inconsistent correlation with human judgements, suggesting that stylistic quality perceived by humans is not captured by simple feature statistics.
Across all evaluation heads, our metric consistently achieves substantially higher correlation with study-derived rankings, demonstrating a stronger alignment with the perceptual criteria governing neural style transfer quality.

\noindent{
\textbf{Generalisation to unseen content and styles.}
The leave-one-method-out (LOMO) protocol above directly reflects the intended use of ASTRA-Score: evaluating a new style transfer method applied to the fixed content and style references in ASTRA-Data. In each split, the evaluated stylisation method is unseen during training, while the benchmark references remain available. We additionally conduct leave-one-content-out (LOCO) and leave-one-style-out (LOSO) experiments as robustness analyses rather than as tests of the intended deployment setting. These experiments examine whether the strong performance of ASTRA-Score depends on having observed every benchmark reference during training. By withholding an entire content or style reference together with all associated stylised results, LOCO and LOSO provide a stricter test of reference dependence and indicate whether the learnt evaluator captures transferable relationships between reference and stylised images rather than simply memorising the specific content and style images in the benchmark.

For LOCO, all stylised images associated with one content reference are held out for testing, while ASTRA-Score is trained on the remaining five contents; this is repeated for all six contents. Table~\ref{tab:metric_correlation_contentwise} reports the results. ASTRA-Score achieves Spearman correlations of 0.745, 0.814, and 0.793 for content, style, and overall preference, respectively, outperforming the strongest baselines by 53.9\% for style and 45.5\% for overall preference. DINO achieves a slightly higher correlation (0.759 vs.\ 0.745), consistent with the strong semantic and structural representations learnt through large-scale self-supervised pretraining. Overall, ASTRA-Score remains strongly aligned with human judgements despite excluding an entire content reference and its associated samples from training.

For LOSO, all stylised images associated with one style reference are similarly held out, with results averaged over the twelve styles (Table~\ref{tab:metric_correlation_stylewise}). ASTRA-Score achieves Spearman correlations of 0.711, 0.700, and 0.658 for content, style, and overall preference, respectively, outperforming the strongest baselines of 0.673, 0.522, and 0.525. Together, the LOCO and LOSO results indicate that ASTRA-Score learns transferable evaluation cues rather than simply memorising the content and style references observed during training.

\noindent\textbf{Robustness across evaluation protocols.}
Taken together, the three protocols examine complementary aspects of ASTRA-Score. LOMO constitutes the primary evaluation and directly reflects its intended use: assessing unseen stylisation methods on the ASTRA benchmark. LOCO and LOSO provide stricter tests by additionally withholding entire content or style references from training. The strong correlations under these settings indicate that ASTRA-Score does not simply memorise reference-specific patterns, but learns relationships between content, style, and stylised outputs that generalise to unseen references. Although such reference-level generalisation extends beyond the primary intended use of ASTRA-Score, the results demonstrate its robustness when applied to content or style references not observed during training.
\begin{table}[htbp]
    \centering
    \caption{Correlation between automatic metrics and study-derived rankings across the evaluated methods.
    Baseline metrics are evaluated using leave-one-style-out (LOSO) evaluation and reported as mean $\pm$ standard deviation.}
    \setlength{\tabcolsep}{3.5pt}
    \begin{tabular}{lcc}
        \toprule
        Metric & Spearman & Kendall \\
        \midrule

        \rowcolor{gray!10}
        \multicolumn{3}{l}{\textbf{Content}} \\
        CFSD~\cite{chung2024style}
            & 0.250 $\pm$ 0.131
            & 0.178 $\pm$ 0.091 \\
        GMSD~\cite{xue2013gradient}
            & 0.391 $\pm$ 0.179
            & 0.280 $\pm$ 0.133 \\
        Qwen~\cite{bai2023qwen}
            & 0.417 $\pm$ 0.191
            & 0.325 $\pm$ 0.143 \\
        FSIM~\cite{zhang2011fsim}
            & 0.435 $\pm$ 0.168
            & 0.312 $\pm$ 0.125 \\
        SIFID~\cite{shaham2019singan}
            & 0.467 $\pm$ 0.152
            & 0.334 $\pm$ 0.114 \\
        VIF~\cite{sheikh2006image}
            & 0.483 $\pm$ 0.109
            & 0.357 $\pm$ 0.084 \\
        MS-SSIM~\cite{wang2003multiscale}
            & 0.510 $\pm$ 0.110
            & 0.369 $\pm$ 0.088 \\
        NMI~\cite{studholme1998normalized}
            & 0.522 $\pm$ 0.135
            & 0.371 $\pm$ 0.098 \\
        SRQE~\cite{chen2022quality}
            & 0.524 $\pm$ 0.099
            & 0.385 $\pm$ 0.079 \\
        CLIP~\cite{radford2021learning}
            & 0.627 $\pm$ 0.085
            & 0.452 $\pm$ 0.066 \\
        ViT~\cite{dosovitskiy2020image}
            & 0.631 $\pm$ 0.102
            & 0.458 $\pm$ 0.086 \\
        LPIPS~\cite{zhang2018unreasonable}
            & 0.642 $\pm$ 0.128
            & 0.475 $\pm$ 0.109 \\
        DINO~\cite{caron2021emerging}
            & 0.673 $\pm$ 0.074
            & 0.493 $\pm$ 0.064 \\
        \textbf{ASTRA-Score Content (ours)} & \textbf{0.711 $\pm$ 0.118} & \textbf{0.534 $\pm$ 0.103} \\

        \midrule
        \rowcolor{gray!10}
        \multicolumn{3}{l}{\textbf{Style}} \\
        CSD~\cite{somepalli2024measuring}
            & 0.168 $\pm$ 0.287
            & 0.116 $\pm$ 0.200 \\
        Qwen~\cite{bai2023qwen}
            & 0.222 $\pm$ 0.174
            & 0.170 $\pm$ 0.132 \\
        CLIP~\cite{radford2021learning}
            & 0.230 $\pm$ 0.152
            & 0.161 $\pm$ 0.109 \\
        SIFID~\cite{shaham2019singan}
            & 0.351 $\pm$ 0.148
            & 0.241 $\pm$ 0.104 \\
        DINO~\cite{caron2021emerging}
            & 0.413 $\pm$ 0.213
            & 0.295 $\pm$ 0.159 \\
        Gram Loss~\cite{gatys2016image}
            & 0.467 $\pm$ 0.304
            & 0.350 $\pm$ 0.215 \\
        ViT~\cite{dosovitskiy2020image}
            & 0.488 $\pm$ 0.198
            & 0.351 $\pm$ 0.157 \\
        SRQE~\cite{chen2022quality}
            & 0.522 $\pm$ 0.198
            & 0.384 $\pm$ 0.146 \\
        \textbf{ASTRA-Score Style (ours)} & \textbf{0.700 $\pm$ 0.143} & \textbf{0.528 $\pm$ 0.130} \\

        \midrule
        \rowcolor{gray!10}
        \multicolumn{3}{l}{\textbf{Overall}} \\
        Qwen~\cite{bai2023qwen}
            & 0.082 $\pm$ 0.177
            & 0.064 $\pm$ 0.134 \\
        ArtFID~\cite{wright2022artfid}
            & 0.192 $\pm$ 0.227
            & 0.135 $\pm$ 0.160 \\
        SRQE~\cite{chen2022quality}
            & 0.525 $\pm$ 0.193
            & 0.377 $\pm$ 0.142 \\
        \textbf{ASTRA-Score Overall (ours)} & \textbf{0.658 $\pm$ 0.231} & \textbf{0.499 $\pm$ 0.198} \\

        \bottomrule
    \end{tabular}
    \label{tab:metric_correlation_stylewise}
\end{table}
\section{Conclusion}
In this paper, we systematically investigate the reliability of automatic evaluation metrics for neural style transfer using a human preference benchmark.
Our results reveal clear limitations of existing metrics, which often fail to align with perceptual judgement, while conventional user studies remain time-consuming, difficult to reproduce, and lack comparability across works.
In this paper, we propose ASTRA, an evaluation mechanism for style transfer. We construct \textbf{ASTRA-Data}, a representative yet tractable benchmark dataset of content–style pairs with pairwise preference annotations, from which we derive ranking-based ground truth for content preservation, style similarity, and overall preference.
We used the user study data to train \textbf{ASTRA-Score}, an evaluation model that predicts preference-aligned scores from content-style-stylisation image triplet, enabling scalable and consistent evaluation across methods: ASTRA-Score provides reliable estimates of stylisation effectiveness on our benchmark image set. Together, ASTRA-Data and ASTRA-Score provide a standardised testbed for benchmarking style transfer methods via their alignment with human rankings.

\newcommand{\etalchar}[1]{$^{#1}$}

\newpage

\clearpage
\appendix
\twocolumn[
\begin{center}
{\huge ASTRA-Score: Focused Perceptual Evaluation of Style Transfer (Appendix)\par}
\end{center}
]

This appendix provides supplementary details and extended analyses to support the main paper. It first describes the construction of ASTRA, including the selection of style images and the generation of stylised results using both learning-based and hand-crafted approaches. It then presents the full user study protocol, covering the training phase and quality control procedures to ensure annotation reliability. Additional analyses are provided to validate the data collection process and to further examine the content–style trade-off across different conditions. Finally, implementation details of the proposed ASTRA-Score evaluation model are reported, including feature extraction, regression head design, and training settings.

\section{Benchmark Data Selection and Generation Details}
This section provides a detailed account of the construction of ASTRA. It outlines the principles guiding the selection of style images, ensuring diversity in artistic characteristics such as colour, texture, structural complexity, and cultural origin. In addition, it describes the implementation details of the LLIP (low-level image processing) methods, which are designed to complement learning-based approaches by approximating styles that are difficult to capture with existing neural models.

\subsection{Style Images Selection}
\label{sec:style image selection}

\textbf{Hokusai (nature, low colour, low detail; Asia, 19th century)}
This iconic woodblock print contains highly recognizable motifs, most notably the sweeping arcs of the crashing wave. It has minimal shading and limited texture, instead relying on strong contour lines and flat regions of solid colour. It has seen use in prior NST evaluation, providing some continuity (original image available at \url{https://commons.wikimedia.org/wiki/File:The_Great_Wave_off_Kanagawa.jpg}).

\textbf{Rembrandt (people, low colour, high detail; Europe, 17th century)}
The chosen image is characterized by strong global contrast and a muted palette. The illuminated areas contain detailed, naturalistic rendering of facial features and fabric, while the background regions are darker and simpler. The combination of high detail, low colour saturation, and straightforward overall composition means that it should be easy to evaluate the effectiveness of stylization while potentially challenging to achieve it (original image available at \url{https://commons.wikimedia.org/wiki/File:Rembrandt,_Self_Portrait_at_the_Age_of_34.jpg}).

\textbf{Chinese Landscape (nature, low colour, high detail; Asia, 19th century)}
This image, in the style of a traditional landscape painting, uses fine brushwork and texture to convey foliage, landscape, and atmospheric depth. The style includes large areas of empty or lightly worked space, which can be difficult for NST methods that assume dense texture. The image is monochromatic, relying strictly on tone and line weight to convey content (original image available at \url{https://commons.wikimedia.org/wiki/File:%E6%98%8E_%E6%B2%88%E5%91%A8_%EF%BC%8C_%E6%96%87%E5%BE%B5%E6%98%8E-%E5%90%88%E7%92%A7%E5%B1%B1%E6%B0%B4%E5%9C%96_%E5%8D%B7-Joint_Landscape_MET_DP235652.jpg}).

\textbf{Bierstadt (nature, high colour, high detail; North America, 19th century)}
The selected landscape is brightly coloured and highly detailed, with dramatic lighting and a strong sense of depth. Although the painting aims for a plausible naturalistic appearance, it remains clearly painterly, but subtly so: methods that depend on crude cues such as severe texture will struggle.
The image provides a foil to the Veronese, which is colourful but less dramatic and offers its own challenges (original image available at \url{https://commons.wikimedia.org/wiki/File:Bierstadt_Albert_On_the_Saco.jpg}).

\textbf{Picasso (people, low colour, low detail; Europe, 20th century)}
Guernica is among the most important and influential works of the 20th century. It contains deliberate distortion of human figures, simplified shapes, and a monochromatic palette. Portions of the image resemble line drawing. This image challenges the proposition that style and content can be cleanly separated (original image available at \url{https://commons.wikimedia.org/wiki/File:Mural_del_%22Guernica%22_de_Picasso.jpg}).

\textbf{Carr (nature, high colour, low detail; North America, 19th century)}
The selected painting uses bold, simplified forms and an unconventional colour scheme. Geometric distortion and broad, smooth regions dominate the composition. We anticipate that texture will be uninformative for this style, while colour and (when available for a given method) distortion of shape will dominate (original image available at \url{https://commons.wikimedia.org/wiki/File:Emily_Carr_-_Trees_in_France.jpg}).

\textbf{Veronese (people, high colour, high detail; Europe, 16th century)}
This Renaissance painting contains clearly defined objects, controlled lighting, and a colourful, harmonious palette. Brush texture is minimal; the rendering style is precise. Like the Bierstadt, this image aims at a realistic aesthetic, yet is clearly painterly (original image available at \url{https://commons.wikimedia.org/wiki/File:Paolo_Veronese_-_The_Marriage_at_Cana_(detail)_-_WGA24859.jpg}).

\textbf{Australian Aboriginal (artificial, high colour, low detail; Australia, 20th century)}
The selected work, a modern exemplar of a traditional style, incorporates motifs such as dots, circles, and sinusoidal lines. These discrete elements do not form continuous textures, making them difficult for texture‑based NST approaches. Success on this style would indicate an ability to reproduce styles with distinct individual elements, such as mosaics (original image available at \url{https://commons.wikimedia.org/wiki/File:Aboriginal-art-503444_960_720.jpg}).

\textbf{Mondrian (artificial, high colour, low detail; Europe, 20th century)}
This image consists of rectangular regions of primary colours separated by black lines. Its deceptive simplicity depends on exact geometry and precise colour placement. We expect many NST methods to struggle with this style; indeed, while reproducing the style algorithmically is feasible, it is not clear how the rectangular blocks can convey concrete subject matter. We include this image as a stress test and as motivation for future work (original image available at \url{https://commons.wikimedia.org/wiki/File:Piet_Mondriaan,_1921_-_Composition_en_rouge,_jaune,_bleu_et_noir.jpg}).

\textbf{Kandinsky (artificial, high colour, high detail; Europe, 20th century)}
The selected painting contains numerous geometric and distinct visual motifs. Although abstract, the elements are flexible enough to be repurposed into representational contexts. The complexity and specificity of these elements make the style difficult to capture (original image available at \url{https://commons.wikimedia.org/wiki/File:Vassily_Kandinsky,_1939_-_Composition_10.jpg}).

\textbf{Ptolemaic Egyptian (people, colourful, low detail; Africa, 350–300 BCE)}
This image features flat regions of colour, clear separation between foreground and background, and stylised depictions of human figures. The perspective and proportions follow historical conventions rather than naturalistic ones. These properties test whether methods can reproduce stylised structure without imposing naturalistic shading or texture (original image available at \url{https://commons.wikimedia.org/wiki/File:P1200377_Louvre_Stele_Ousirour_detail_D_N2699_rwk.jpg}).

\textbf{Gris (artificial, low colour, low detail; Europe, 20th century)}
The fragmented shapes, overlapping planes, and occasional symbolic elements in this work make it supremely challenging. The central violin is depicted partly by shape and partly by line drawing. Although the palette is limited, the spatial organization is complex. Balancing content preservation with stylistic transformation will be a severe challenge (original image available at \url{https://commons.wikimedia.org/wiki/File:Juan_Gris_-_Komposition_mit_Violine.jpeg}).

\subsection{Details of LLIP Method Construction}
\label{sec:Details of LLIP Method Construction}

The LLIP (low-level image processing) category represents a set of hand-crafted approaches designed to approximate diverse artistic styles.
Given the absence of a single model capable of covering all styles, we construct this set using a combination of existing methods, hybrid pipelines, and custom-designed procedures.

\subsubsection{Design Strategies}

We adopt four main strategies to construct LLIP stylisations:

\begin{itemize}
    \item \textbf{Style-aligned method selection.}
    Existing methods are selected when their visual characteristics align well with the target style (e.g., Kandinsky, Bierstadt).

    \item \textbf{Method combination.}
    Multiple methods are combined, either sequentially or by averaging their outputs, to better approximate complex styles (e.g., Rembrandt, Chinese landscape, Veronese).

    \item \textbf{Multi-style methods.}
    Some approaches are capable of representing multiple styles and are directly applied (e.g., image analogies for Aboriginal art and The Great Wave).

    \item \textbf{Custom  pipelines.}
    For styles not well supported by existing methods, custom pipelines are designed (e.g., Mondrian, Carr, Picasso, Ptolemaic Egyptian).
\end{itemize}

\subsubsection{Style-Specific Implementations}

\paragraph{Aboriginal and Great Wave.}
We employ image analogies~\cite{jacobs2001image} as the base method.
For Great Wave, this is further combined with colour transfer using distribution matching~\cite{pitie2007automated}.

\paragraph{Rembrandt.}
We combine bilateral filtering~\cite{tomasi1998bilateral} with a custom colour mapping procedure based on look-up tables (LUT).
We applied histogram matching between grayscale versions of the content and style images.
For each intensity level, the average RGB colour is computed from the style image to construct a mapping, which is then applied to the content image.

\paragraph{Kandinsky.}
We use a structured abstraction framework~\cite{faraj2017generic} to produce simplified geometric representations consistent with the style.

\paragraph{Bierstadt.}
We apply the brushify filter from G'MIC~\cite{tschumperle2023g}.
For images containing foreground objects, the foreground is processed using brushify, while the background is rendered using a watercolour model~\cite{rosin2017watercolour}.

\paragraph{Veronese.}
We compute the average of multiple stylisations, including:
(1) brushify filtering~\cite{tschumperle2023g},
(2) $L_0$ smoothing~\cite{xu2011image},
(3) outputs from the Waterlogue application, and
(4) adaptive morphological filtering~\cite{blusseau2022adaptive}.

\paragraph{Chinese Landscape.}
We combine brush-based rendering~\cite{tschumperle2023g} with sketch extraction~\cite{lu2012combining}, and average the resulting outputs.

\paragraph{Ptolemaic Egyptian.}
A custom pipeline is designed as follows: the image is first blurred and segmented into a fixed number of colour layers based on basic colour categories~\cite{berlin}.
Textures are added to each layer, and line drawings extracted using coherent line detection~\cite{kang2007coherent} are overlaid.
The process is applied to both the original and intensity-equalised images, and the results are combined.
Background regions are recoloured separately using a reduced palette.

\paragraph{Picasso.}
A stylised abstraction is constructed by converting the image to grayscale, quantising intensities, and applying median filtering.
Line drawings~\cite{kang2007coherent} are overlaid, followed by the addition of structured noise and grid-like patterns.

\paragraph{Mondrian.}
The image is downscaled and each pixel is mapped to a restricted colour palette.
Rectangular regions are iteratively extracted and expanded to form block structures, with boundary lines extended to create a grid-like composition.

\paragraph{Carr.}
We segment the content image with SLIC0, generating medium-large regions that approximately conform to object boundaries. Next, we assign a random painterly texture to each, recolouring the texture to match the average content colour within each region. Note that this emphasizes shape abstraction and not colour; in particular, it does not use the unusual colour distribution characteristic of Carr's style.

\begin{figure}[t]
    \centering
    \includegraphics[width=\linewidth]{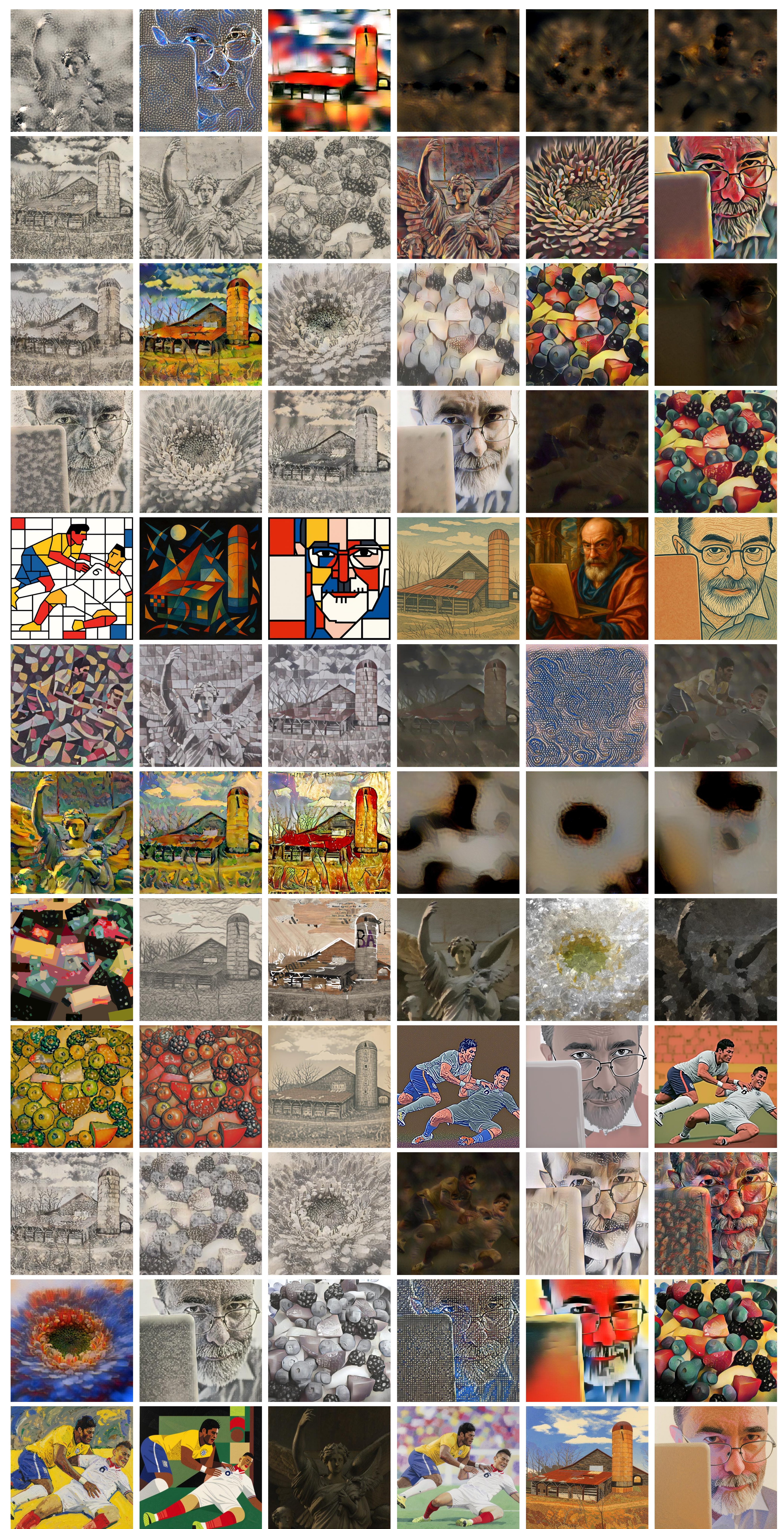}
    \caption{Top-3 and bottom-3 stylised results for each method ranked by overall human preference scores. In each row, the left three images show the top-3 results and the right three show the bottom-3 results. Methods are listed in alphabetical order: AAMS, AdaAttN, AdaIN, ArtFlow, ChatGPT, NST-Ghiasi, Gatys, LLIP, OmniStyle, SANET, StyTR-2, and USO.}

    \label{fig:top3_bottom3_each_method}
\end{figure}

\section{User Study Protocol}
This section describes the design of the user study and the procedures used to ensure the reliability and consistency of the collected annotations. It first outlines the training phase, which establishes a shared understanding of the evaluation criteria among participants. After introducing the three evaluation criteria, few illustrative examples was presented with unambiguous answers (distinct from the main survey), with further details provided below.It then details the quality control mechanisms applied during the study, including attention checks and fatigue mitigation strategies. Together, these components ensure that the resulting annotations are both interpretable and robust for subsequent analysis.
\subsection{Training Phase}
\label{sec:user_study_training}

Before entering the main study, participants completed a mandatory training phase intended to provide a consistent understanding of the evaluation criteria. This phase consisted of an introductory tutorial explaining the key concepts of \emph{content}, \emph{style}, and \emph{style transfer}, supported by both textual descriptions and visual examples.

Participants were required to complete a set of training questions during this phase. Immediate feedback was provided after each response, and participants were required to answer correctly before proceeding. This process gave each participant a minimum level of understanding prior to the evaluation tasks.

\paragraph{Content.}
The \emph{content} of an image refers to the scene being represented, regardless of stylistic rendering, as shown in Fig.~\ref{fig:tutorial_content}. Participants were instructed to consider all scene elements, including both foreground and background components. While stylistic variations (e.g., blur or abstraction) may affect how these elements are depicted, a style should not introduce new semantic objects. Any such deviation was described as reducing content faithfulness.

\begin{figure}[t]
    \centering
    \includegraphics[width=0.9\linewidth]{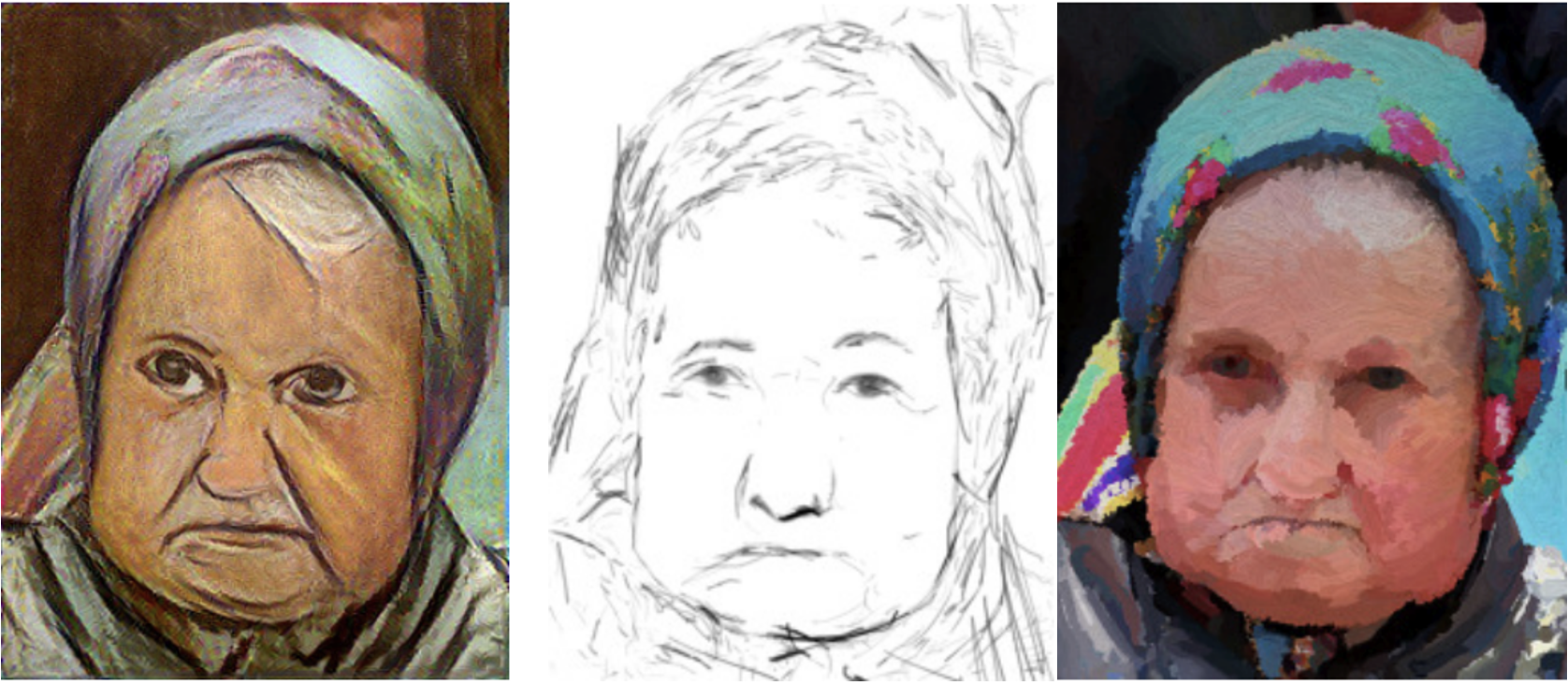}
    \caption{Illustration of content consistency: the same scene rendered in different styles.}

    \label{fig:tutorial_content}
\end{figure}

\paragraph{Style.}
The \emph{style} of an image refers to the artistic characteristics that define its visual appearance, including medium, brushstrokes, texture, and colour usage, as shown in Fig.~\ref{fig:tutorial_style}. Participants were informed that style may include colour, particularly when constrained by the medium (e.g., pen-and-ink). However, in some cases colour may also depend on the underlying content rather than stylistic intent, and should therefore be interpreted with care.

\begin{figure}[t]
    \centering
    \includegraphics[width=0.9\linewidth]{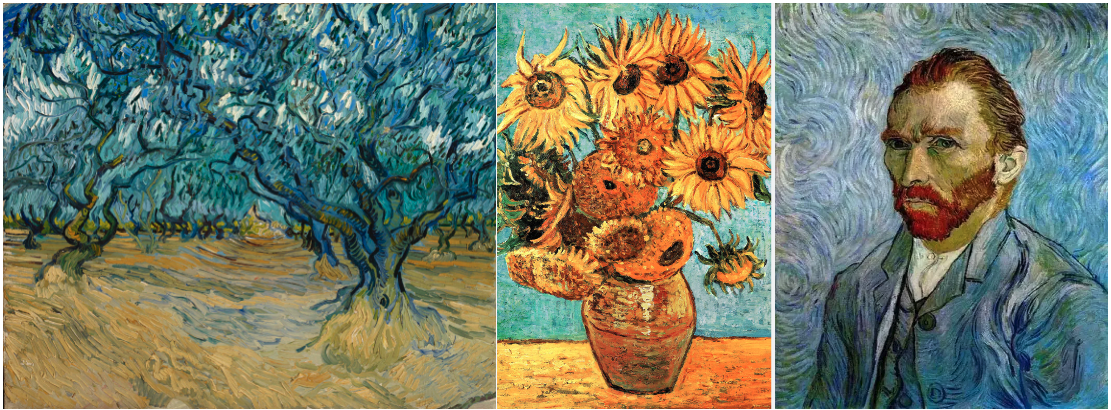}
    \caption{Illustration of style consistency: different contents rendered in the same style.}
    \label{fig:tutorial_style}
\end{figure}

\paragraph{Style Transfer.}
\emph{Style transfer} was defined as the process of recreating the content of one image using the style of another, as shown in Fig.~\ref{fig:tutorial_stylisation}. Participants were instructed to evaluate stylised results based on how well they balance content preservation and style similarity. In particular, they were encouraged to consider an ideal stylisation that faithfully preserves the content while accurately reflecting the target style. Participants were explicitly advised not to select images solely based on subjective preference, but to judge overall quality as a combination of content and style.

\paragraph{Illustrative Examples}
 After introducing the three evaluation dimensions—content preservation, style fidelity, and overall preference—the training phase presents few illustrative examples with clearly distinguishable correct answers, which are distinct from those used in the main survey. For each example, participants are required to make a selection and are immediately provided with feedback indicating whether their choice is correct, along with a brief explanation. Participants must answer each example correctly before proceeding to the next, ensuring that they fully understand the evaluation criteria before entering the main study; examples are shown in Fig.~\ref{fig:Illustrative Examples}.

\subsection{Quality Control}
\label{sec:user_study_quality}
To ensure the reliability of the collected responses, we applied quality control mechanisms during the main study, detailed as follows.

\paragraph{Attention Checks.}
The study included dedicated attention-check trials with predefined correct answers for the \emph{content} and \emph{style} dimensions. These trials were interleaved with regular evaluation tasks to identify inattentive participants.
In Stage~1, each participant completed six attention-check trials. A response was considered correct only if both the content and style questions were answered correctly. Participants were included in the analysis only if they answered at least five out of six attention-check trials correctly.
In Stage~2, each participant completed four attention-check trials under the same criteria. Participants were retained only if they answered at least three out of four trials correctly.

\paragraph{Break and Fatigue Control.}
To mitigate fatigue effects in long sessions, we included a mandatory timed break at the midpoint of the study. The break cannot be skipped and the study resumes automatically after a fixed interval, reducing cognitive load accumulation and maintaining response reliability.

\leavevmode\paragraph{Response Filtering.}
After applying the attention-check criteria, 264 of the 407 submitted responses were retained for analysis. To further examine whether the filtering procedure identified low-effort responses, we compared response times between retained and rejected submissions. The median response time per question was 4.526\,s for rejected submissions,
compared with 8.573\,s for retained submissions. Thus, rejected participants responded in approximately half the time of retained participants, providing additional evidence that the attention-check criteria identified inattentive or low-effort responses.

\begin{figure}[t]
    \centering
    \includegraphics[width=0.9\linewidth]{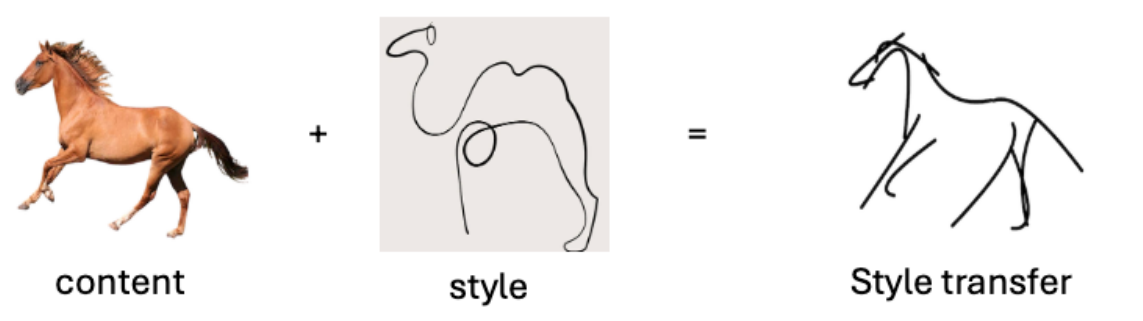}
    \caption{Example of style transfer: content from one image rendered in the style of another.}
    \label{fig:tutorial_stylisation}
\end{figure}

\begin{figure}[t]
  \centering
  \includegraphics[width=\linewidth]{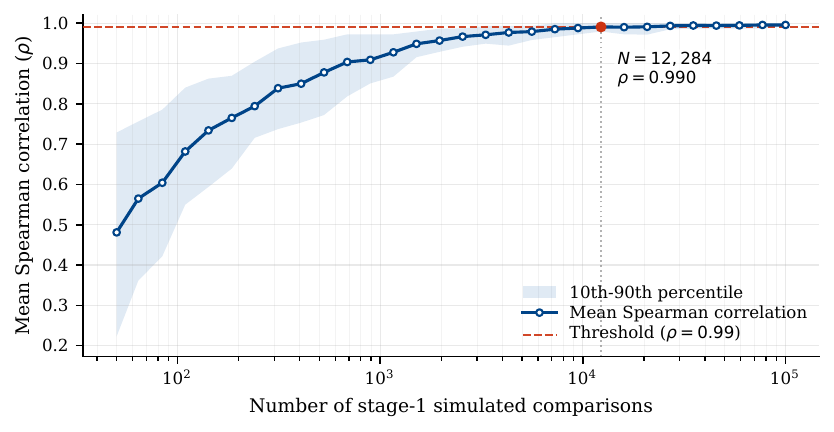}
  \caption{
  Ranking stability under simulated within-group comparisons (Stage~1).
  The Spearman correlation between the recovered ranking and the assumed ground-truth ranking increases as the number of comparisons grows.
  The curve rapidly converges and exceeds $\rho=0.99$ at approximately $1.2\times10^4$ comparisons, indicating that the estimated method ranking becomes highly stable beyond this point.
  Each point represents the mean correlation over multiple simulation runs.
  }
  \label{fig:ranking_stability_stage1}
\end{figure}

\begin{figure}[t]
\centering
    \begin{subfigure}{\linewidth}
        \centering
        \includegraphics[width=\linewidth]{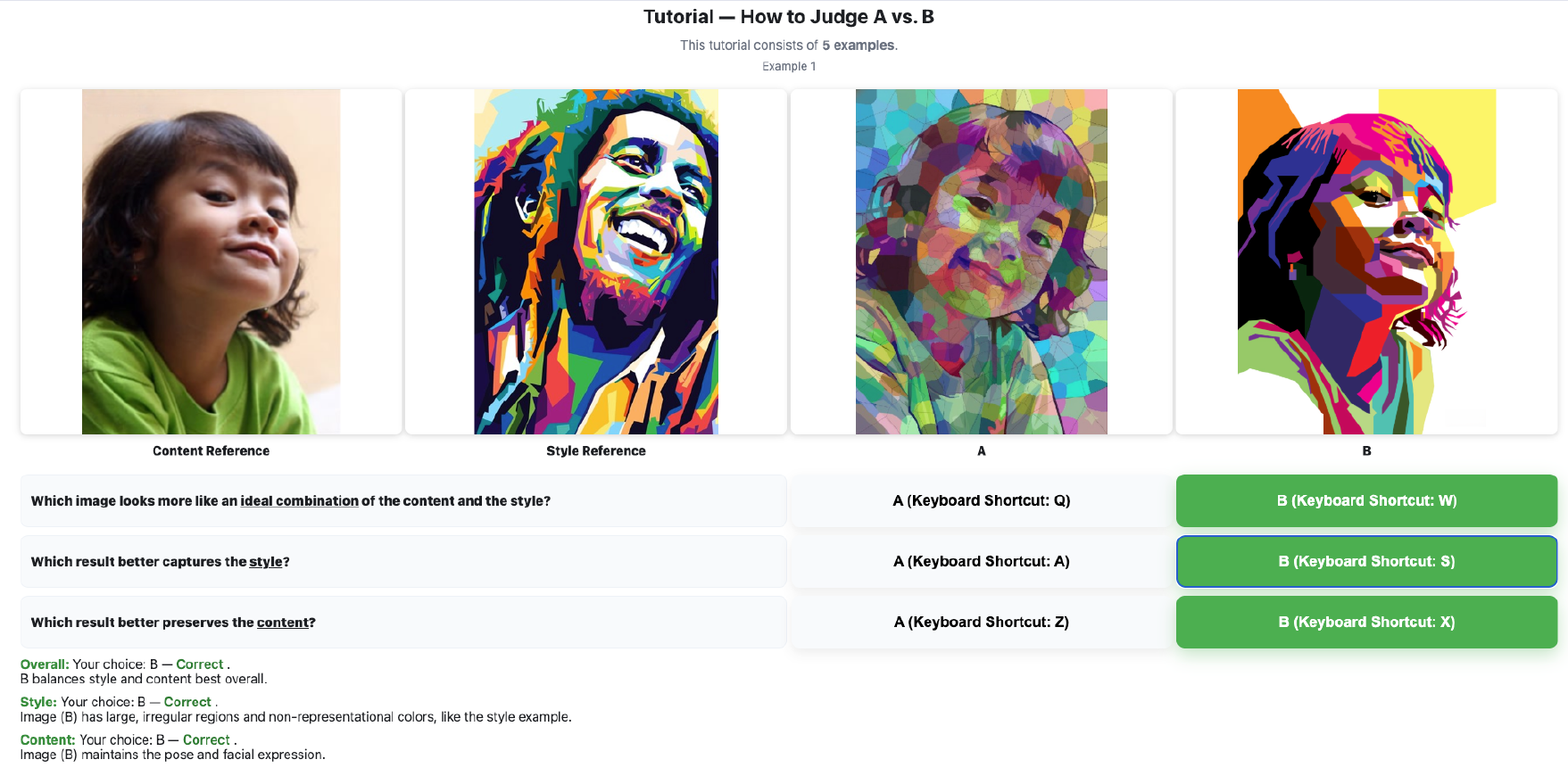}
        \caption{Stage~1 illustrative example.}

    \end{subfigure}
    \vspace{4pt}
    \begin{subfigure}{\linewidth}
        \centering
        \includegraphics[width=\linewidth]{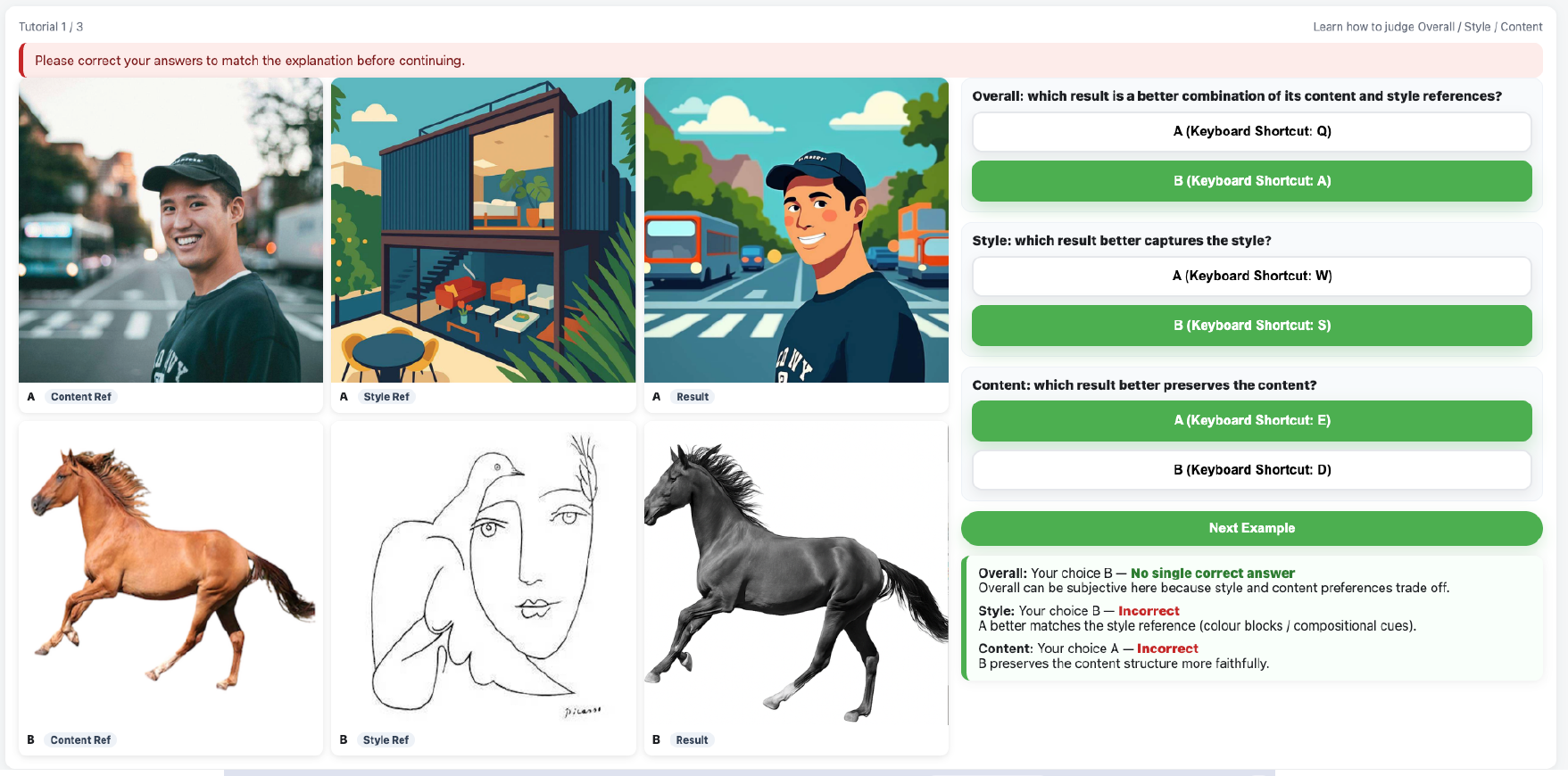}
        \caption{Stage~2 illustrative example.}

    \end{subfigure}
    \caption{
        Examples of the two illustrative examples used in our user study.
        }

    \label{fig:Illustrative Examples}
\end{figure}

\subsection{Annotation Consistency}
\label{app:annotation_consistency}

To quantify the consistency of the collected preference annotations, we randomly selected a subset of comparisons from each stage and obtained duplicate annotations from additional independent annotators. The duplicated comparisons followed the same evaluation protocol as the original study, with annotators independently judging content preservation, style similarity, and overall style-transfer preference.

We measured agreement between the original and additional annotations using Cohen's $\kappa$. For Stage~1, the resulting $\kappa$ values were 0.663 for content preservation, 0.547 for style similarity, and 0.547 for overall preference. For Stage~2, the corresponding values were 0.515, 0.389, and 0.416, respectively. The higher agreement observed in Stage~1 is consistent with its controlled design, where both stylised results share the same content-style references. In contrast, Stage~2 requires comparisons across different content-style pairs and therefore involves a more demanding perceptual judgement.

Overall, these results indicate meaningful but task-dependent annotation consistency despite the subjective nature of style-transfer evaluation. Participants were recruited internationally through online platforms under an approved ethics protocol. The present study was not designed to analyse demographic or cultural differences in perceptual preference; investigating the stability of style-transfer preferences across different cultural or user groups therefore remains an important direction for future work.

\section{Ranking Stability Simulation}
\label{sec:simulation_appendix}
Before conducting the user study,
we analysed the required number of pairwise comparisons from both theoretical and empirical perspectives.
Spectral ranking theory~\cite{negahban2017rank,chen2015spectral} suggests that reliable ranking estimation under models such as Bradley-Terry-Luce~\cite{bradley1952rank,luce1959individual} scales on the order of $\Theta(n\log n)$ comparisons.
Using natural logarithm, this corresponds to approximately $12\ln 12 \approx 30$ comparisons per group for Stage~1 and $864\ln 864 \approx 5.8\times10^3$ comparisons for Stage~2.

These values provided theoretical reference scales rather than exact thresholds, so we further performed simulation analyses under our two-stage design to empirically estimate the number of comparisons needed for stable ranking recovery.
\paragraph{Stage~1: Within-group ranking stability.}
Stage~1 evaluates style transfer methods under identical content-style conditions.
To estimate the number of comparisons required to obtain reliable method preferences, we simulated pairwise comparison processes assuming the existence of an underlying ground-truth ranking among methods.
Each simulated comparison followed the ground-truth ordering with stochastic noise to account for potential human inconsistency.
The simulated outcomes were aggregated using the \textbf{Rank Centrality} algorithm~\cite{negahban2017rank}, a Markov-chain-based estimator for ranking from pairwise comparisons, see more details in Section~\ref{sec:groundtruth_establishment}.
Ranking stability was measured using the Spearman correlation~\cite{spearman1961proof} between the recovered ranking and the assumed ground-truth ranking.

As shown in Fig.~\ref{fig:ranking_stability_stage1}, the recovered ranking rapidly stabilises as the number of comparisons increases and exceeds a correlation of $\rho=0.99$ at approximately $1.2\times10^4$ comparisons.
In our actual user study, Stage~1 collected \textbf{17,568 votes from 122 participants}, which lies well within the regime where the ranking estimation is highly stable.

\paragraph{Stage~2: Global rankings.}
While Stage~1 provides reliable rankings within each content-style group, it does not establish relative ordering across different groups.
Stage~2 acquires cross-group comparisons between stylised images belonging to different content-style pairs in order to construct a global ranking across all items.

To estimate the number of comparisons required for global ranking, we simulated a synthetic environment in which each item was assigned a latent score determined by method, content, and style effects with additional interaction noise.
The combined comparison data are aggregated using Rank Centrality to estimate the global ranking.
\begin{figure}[t]
\centering
\includegraphics[width=\linewidth]{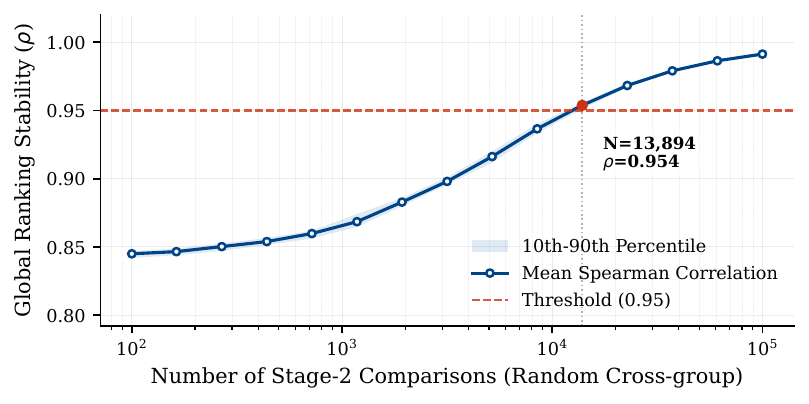}
\caption{
Global ranking recovery under simulated cross-group comparisons (Stage~2).
Additional comparisons between items belonging to different content-style pairs gradually improve the accuracy of the recovered global ranking.
The Spearman correlation steadily increases as more cross-group comparisons are introduced, demonstrating that sufficient Stage~2 comparisons enable reliable global ranking estimation.
}

\label{fig:ranking_stability_stage2}
\end{figure}
As shown in Fig.~\ref{fig:ranking_stability_stage2}, the accuracy of the recovered ranking improves steadily as the number of cross-group comparisons increases.
The Spearman correlation reaches approximately $0.95$ when the number of Stage~2 comparisons approaches $1.4\times10^4$ and continues to increase gradually thereafter.
In practice, Stage~2 collected \textbf{17,040 votes from 142 participants}, which lies within the regime where the recovered ranking is already highly correlated with the ground truth.

These analyses provide empirical evidence that the scale of our two-stage user study is sufficient to support reliable preference modelling and global ranking estimation.

{\color{black}
\subsection{Comparison Graph and Rank Centrality Diagnostics}
\label{sec:graph_diagnostics}

\paragraph{Empirical graph connectivity.}
We represent each of the 864 stylised images as a node and add an undirected edge whenever two images are compared at least once. Stage~1 comparisons are restricted to methods sharing the same content-style pair and therefore form 72 disconnected components. Stage~2 was specifically introduced to connect these local components
through cross-pair comparisons. After combining the two stages, the empirical comparison graph contains a single connected component covering all 864 nodes, with an average node degree of approximately 45.4. Thus, the comparison graph used for Rank Centrality is connected, and the resulting Markov chain admits a global stationary ranking.

\paragraph{Connectivity under Stage~2 sampling.}
The connectivity of the 72 content-style groups can also be analysed under the random Stage~2 sampling procedure. For any subset of $k$ groups, disconnection requires that none of the Stage~2 comparisons
cross the cut between that subset and the remaining $72-k$ groups. Using a union bound over all possible disconnected subsets, the probability of a disconnected group-level graph is bounded by
\begin{equation}
\Pr(\text{disconnected})
\leq
\sum_{k=1}^{36}
\binom{72}{k}
\left(
1-\frac{k(72-k)}{\binom{72}{2}}
\right)^{17040}
\approx 2.4\times10^{-207}.
\end{equation}
We additionally simulated the Stage~2 sampling process $10^6$ times
and did not observe a disconnected graph in any simulation.

\paragraph{Spectral gap.}
We compute the spectral gap of the Rank Centrality transition matrix for each evaluation criterion. The resulting gaps are 0.0199 for content preservation, 0.0421 for style similarity, and 0.1000 for overall preference. All three transition matrices therefore have non-zero empirical spectral gaps. In conjunction with the connectivity
analysis and the ranking-recovery experiments, these results provide additional diagnostics of the Markov chains underlying the estimated preference rankings.
}

\section{ASTRA Analysis}
This section provides additional analyses to validate the reliability of the collected data and to further examine the behavioural patterns underlying human preference. It first analyses the exposure statistics of content and style references across methods to verify that the sampling process is approximately balanced and does not introduce systematic bias. It then extends the investigation of the content–style trade-off by presenting per-style and per-content analyses, offering a more detailed view of how preference patterns vary across different conditions. These analyses provide further evidence supporting the robustness of the benchmark and the style-dependent nature of the observed trade-offs.
\subsection{Exposure Statistics of Content and Style References}
\label{app:exposure}

\begin{table*}[t]
\centering
\small
\caption{Exposure counts between reference styles and style transfer methods in the Stage~1 user study.
Each entry reports the number of pairwise comparisons in which a given method appears under the corresponding style reference.}

\label{tab:stage1_style_exposure}
\setlength{\tabcolsep}{3.5pt}
\begin{tabular}{lcccccccccccc}
\toprule
Style & AAMS & AdaAttN & AdaIN & ArtFlow & ChatGPT & Gatys & LLIP & OmniStyle & SANET & StyTR-2 & USO & NST-Ghiasi \\
\midrule
Australian Aboriginal & 236 & 288 & 230 & 276 & 177 & 238 & 255 & 228 & 268 & 213 & 276 & 241 \\
Bierstadt & 229 & 249 & 280 & 225 & 248 & 281 & 201 & 241 & 276 & 263 & 248 & 227 \\
Carr & 225 & 243 & 266 & 258 & 254 & 274 & 223 & 310 & 230 & 217 & 237 & 195 \\
Hokusai & 215 & 268 & 321 & 272 & 221 & 256 & 232 & 263 & 209 & 221 & 214 & 232 \\
Gris & 246 & 224 & 240 & 231 & 322 & 257 & 271 & 210 & 190 & 268 & 204 & 257 \\
Picasso & 236 & 251 & 291 & 225 & 268 & 251 & 212 & 213 & 290 & 253 & 230 & 208 \\
Ptolemaic Egyptian & 244 & 241 & 237 & 264 & 250 & 234 & 250 & 240 & 214 & 288 & 223 & 235 \\
Veronese & 262 & 213 & 231 & 257 & 194 & 270 & 275 & 211 & 260 & 285 & 204 & 266 \\
Mondrian & 215 & 263 & 212 & 224 & 256 & 223 & 229 & 228 & 241 & 277 & 309 & 261 \\
Rembrandt & 271 & 217 & 251 & 237 & 189 & 244 & 234 & 248 & 255 & 280 & 301 & 195 \\
Kandinsky & 226 & 269 & 252 & 254 & 259 & 248 & 264 & 267 & 220 & 234 & 224 & 199 \\
Chinese Landscape & 281 & 257 & 215 & 237 & 198 & 276 & 223 & 274 & 250 & 235 & 249 & 219 \\
\bottomrule
\end{tabular}
\end{table*}

\begin{table*}[t]
\centering
\small
\caption{Exposure counts between content images and style transfer methods in the Stage~1 user study.
Each entry indicates the number of pairwise comparisons in which a given method appears for the corresponding content image.}

\label{tab:stage1_content_exposure}
\begin{tabular}{lcccccccccccc}
\toprule
Content & AAMS & AdaAttN & AdaIN & ArtFlow & ChatGPT & Gatys & LLIP & OmniStyle & SANET & StyTR-2 & USO & NST-Ghiasi \\
\midrule
angel & 516 & 493 & 544 & 469 & 500 & 491 & 480 & 453 & 512 & 522 & 425 & 437 \\
athletes & 509 & 522 & 479 & 474 & 432 & 428 & 515 & 496 & 486 & 543 & 548 & 402 \\
barn & 389 & 520 & 549 & 498 & 497 & 494 & 437 & 473 & 525 & 479 & 510 & 489 \\
berries & 507 & 506 & 459 & 562 & 462 & 523 & 443 & 516 & 458 & 496 & 431 & 491 \\
daisy & 506 & 494 & 476 & 489 & 468 & 549 & 496 & 521 & 469 & 485 & 489 & 456 \\
mac & 459 & 448 & 519 & 468 & 477 & 567 & 498 & 474 & 453 & 509 & 516 & 460 \\
\bottomrule
\end{tabular}
\end{table*}

\begin{table*}[t]
\centering
\small
\caption{Exposure counts between style references and methods in Stage~2 comparisons.}

\label{tab:style_exposure}
\begin{tabular}{lcccccccccccc}
\toprule
Style & AAMS & AdaAttN & AdaIN & ArtFlow & ChatGPT & Gatys & LLIP & OmniStyle & SANET & StyTR-2 & USO & NST-Ghiasi \\
\midrule
Australian Aboriginal & 214 & 254 & 264 & 246 & 232 & 237 & 254 & 228 & 229 & 202 & 250 & 269 \\
Bierstadt & 249 & 231 & 245 & 231 & 245 & 255 & 241 & 234 & 251 & 209 & 265 & 247 \\
Carr & 241 & 219 & 229 & 219 & 223 & 239 & 223 & 243 & 235 & 228 & 225 & 251 \\
Hokusai & 238 & 248 & 233 & 223 & 210 & 219 & 239 & 237 & 236 & 252 & 217 & 250 \\
Gris & 215 & 245 & 240 & 251 & 243 & 242 & 234 & 255 & 259 & 221 & 239 & 269 \\
Picasso & 209 & 237 & 241 & 228 & 232 & 237 & 252 & 223 & 222 & 248 & 257 & 241 \\
Ptolemaic Egyptian & 226 & 235 & 240 & 252 & 238 & 239 & 229 & 224 & 223 & 213 & 241 & 273 \\
Veronese & 233 & 238 & 262 & 247 & 220 & 233 & 260 & 267 & 224 & 247 & 238 & 228 \\
Mondrian & 239 & 211 & 203 & 231 & 226 & 228 & 255 & 231 & 242 & 236 & 231 & 237 \\
Rembrandt & 201 & 240 & 231 & 253 & 241 & 249 & 219 & 244 & 228 & 231 & 258 & 220 \\
Kandinsky & 226 & 235 & 236 & 222 & 242 & 229 & 249 & 250 & 249 & 237 & 249 & 223 \\
Chinese Landscape & 213 & 257 & 236 & 253 & 222 & 235 & 230 & 255 & 219 & 218 & 217 & 264 \\
\bottomrule
\end{tabular}
\end{table*}

\begin{table*}[t]
\centering
\small
\caption{Exposure counts between content images and methods in Stage~2 comparisons.}

\label{tab:content_exposure}
\begin{tabular}{lcccccccccccc}
\toprule
Content & AAMS & AdaAttN & AdaIN & ArtFlow & ChatGPT & Gatys & LLIP & OmniStyle & SANET & StyTR-2 & USO & NST-Ghiasi \\
\midrule
Angel & 442 & 477 & 490 & 528 & 468 & 505 & 473 & 471 & 454 & 465 & 480 & 490 \\
Athletes & 456 & 493 & 461 & 441 & 503 & 502 & 494 & 504 & 444 & 452 & 482 & 520 \\
Barn & 473 & 508 & 478 & 500 & 486 & 470 & 510 & 454 & 497 & 452 & 480 & 466 \\
Berries & 446 & 437 & 492 & 470 & 423 & 445 & 472 & 474 & 438 & 460 & 459 & 519 \\
Daisy & 441 & 476 & 503 & 475 & 419 & 459 & 466 & 504 & 491 & 456 & 515 & 481 \\
Mac & 446 & 459 & 436 & 442 & 475 & 461 & 470 & 484 & 493 & 457 & 471 & 496 \\
\bottomrule
\end{tabular}
\end{table*}

To verify that the preference score estimation is not dominated by specific methods, we analyse the exposure frequency between references and methods.
Tab.~\ref{tab:style_exposure} and Tab.~\ref{tab:content_exposure} report the number of times each style or content reference appears together with each method across all Stage~2 comparisons.

These statistics demonstrate that the sampling distribution is approximately balanced.
For style references, each method appears roughly $200$–$270$ times per style.
For content references, each method appears approximately $420$–$530$ times per content image.
This balanced exposure ensures that method-specific biases are largely averaged out when aggregating pairwise judgements into Rank Centrality scores.

\subsection{Additional Analysis of Content-Style Trade-off}
\label{sec:pareto_curve}

To further investigate the content-style trade-off, we extend the analysis presented in the main paper by examining the preference behaviour under all individual style and content conditions.

\paragraph{Per-style Analysis.}
For each style reference, we compute the style and content win rates for all methods based on Stage~1 pairwise comparisons. Each method is then represented as a point in the content-style plane.

Fig.~\ref{fig:pareto_all_styles} presents the resulting distributions across a diverse set of styles; each subfigure corresponds to one style reference.

\begin{figure*}[t]
    \centering

    \includegraphics[width=0.32\linewidth]{pics/pareto_curve/pareto_Piet_Mondriaan_1921_-_Composition_en_rouge_jaune_bleu_et_noir-CROP-512_png.pdf}
    \includegraphics[width=0.32\linewidth]{pics/pareto_curve/pareto_Vassily_Kandinsky,_1939_-_Composition_10-CROP-512_png.pdf}
    \includegraphics[width=0.32\linewidth]{pics/pareto_curve/pareto_Rembrandt,_Self_Portrait_at_the_Age_of_34-CROP-512_png.pdf}
    \vspace{4pt}
    \includegraphics[width=0.32\linewidth]{pics/pareto_curve/pareto_Bierstadt_Albert_On_the_Saco-CROP-512_png.pdf}
    \includegraphics[width=0.32\linewidth]{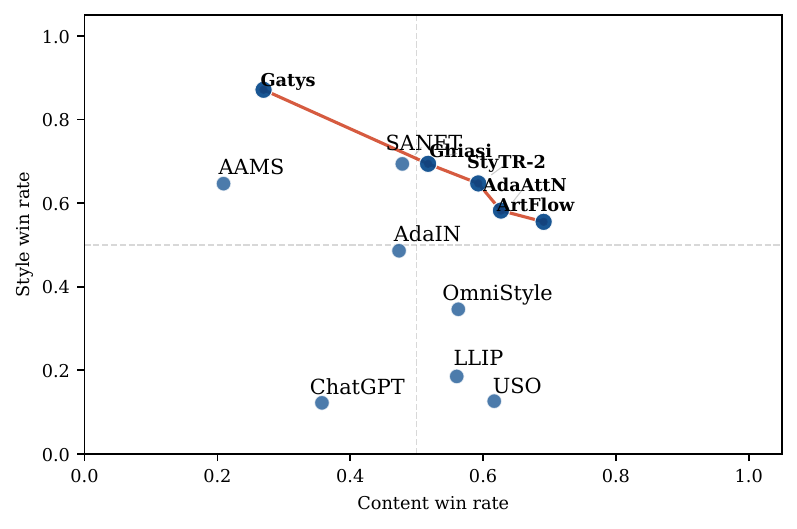}
    \includegraphics[width=0.32\linewidth]{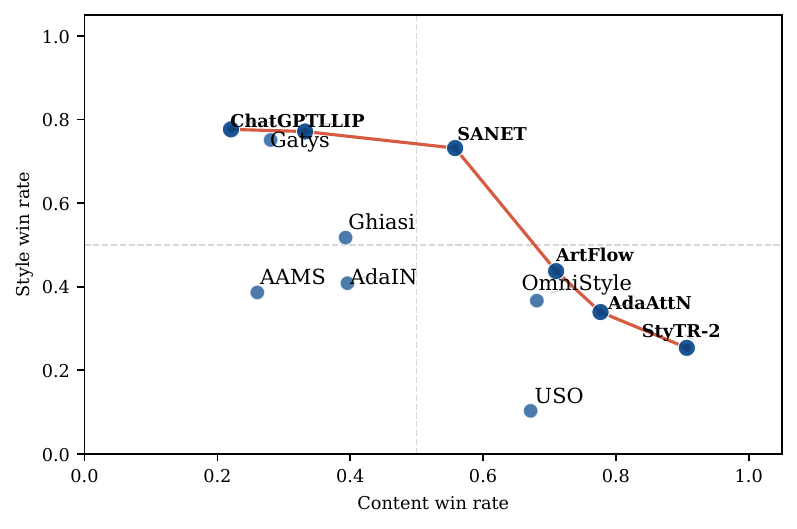}
    \vspace{4pt}
    \includegraphics[width=0.32\linewidth]{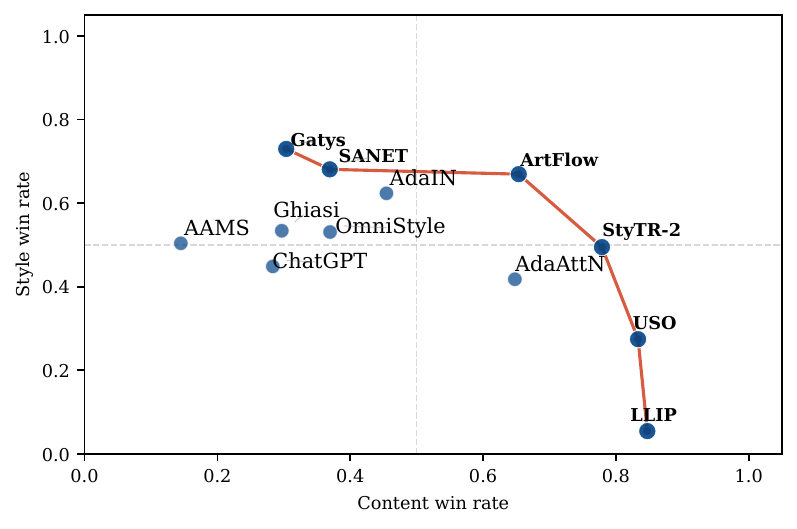}
    \includegraphics[width=0.32\linewidth]{pics/pareto_curve/pareto_Vassily_Kandinsky,_1939_-_Composition_10-CROP-512_png.pdf}
    \includegraphics[width=0.32\linewidth]{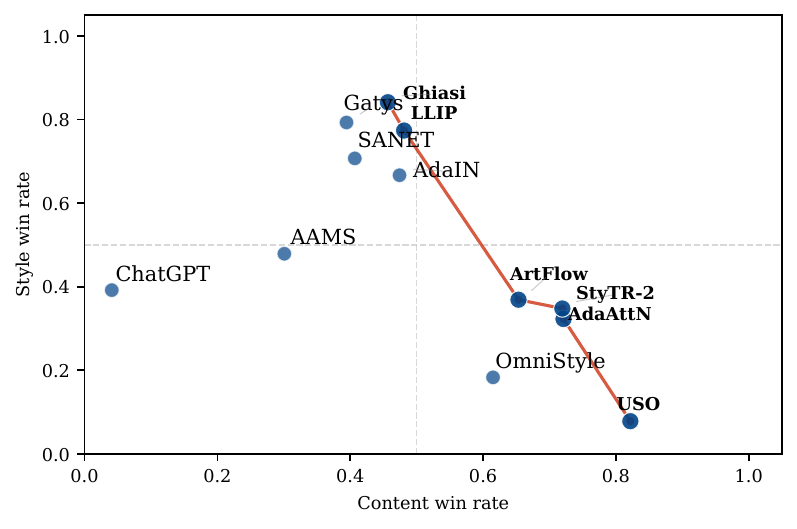}
    \vspace{4pt}
    \includegraphics[width=0.32\linewidth]{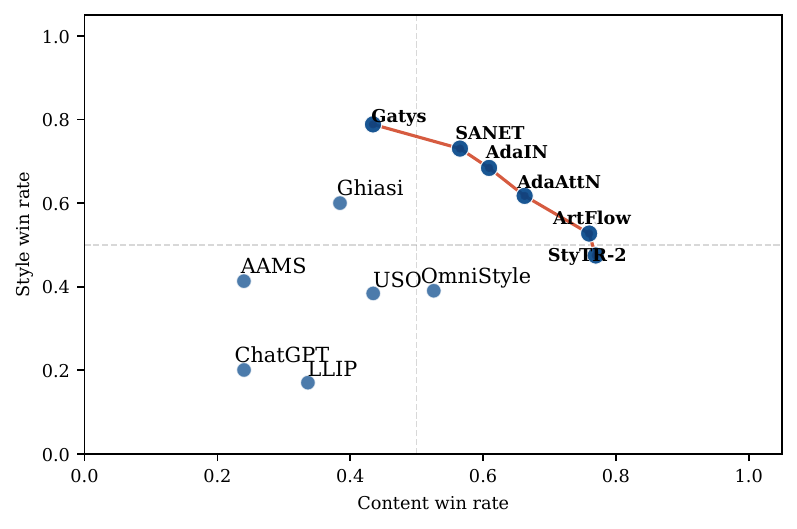}
    \includegraphics[width=0.32\linewidth]{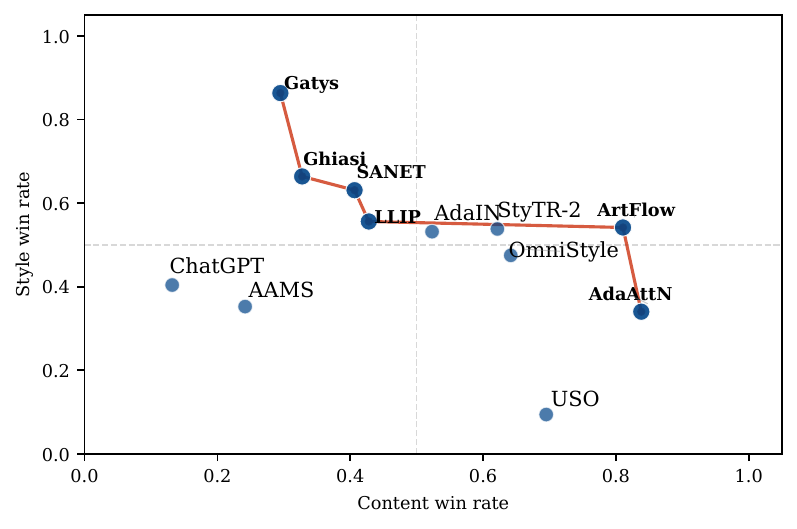}
    \includegraphics[width=0.32\linewidth]{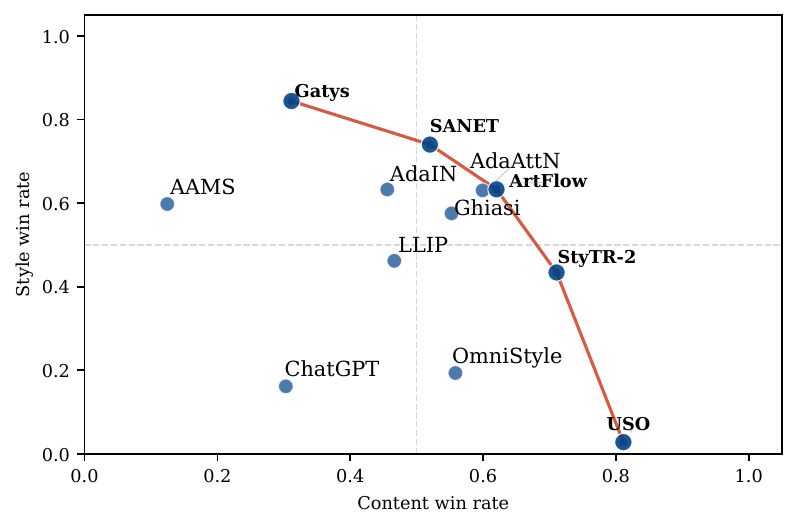}

    \caption{
    Content-style preference across different style references.
    Each subplot corresponds to a specific style, arranged from left to right and top to bottom as:
    Mondrian, Carr, Chinese Landscape, Kandinsky, Bierstadt, Rembrandt, Picasso, Australian Aboriginal, Hokusai, Ptolemaic Egyptian, Gris, and Veronese.
    Each point represents a stylisation method positioned by its content and style win rates.
    The red curve indicates the Pareto frontier of the two objectives, which highlights the optimal trade-off boundary between the two criteria.
    The relationship between content preservation and style similarity varies significantly across styles, highlighting strong style-dependent behaviour.
    }

    \label{fig:pareto_all_styles}
\end{figure*}

\paragraph{Observations.}
Some consistent patterns can be observed across styles:

\begin{itemize}
    \item \textbf{Abstract styles} (e.g., Mondrian, Kandinsky) exhibit a clear inverse relationship between content and style preferences. Methods that achieve stronger stylistic effects often introduce structural distortions, leading to reduced content fidelity.

    \item Only in Ptolemaic Egyptian case, a positive correlation between content and style preferences can be observed.

    \item In other styles, no clear trend emerges, indicating that the trade-off is not universal but highly dependent on the specific style properties.
\end{itemize}

\paragraph{Per-content Analysis.}
To further examine whether the observed preferences depends on the input content, we group the results by content images and compute the corresponding content and style win rates for each method.

Fig.~\ref{fig:pareto_all_contents} presents the resulting distributions for all content images. Each subplot corresponds to one content instance.

\begin{figure*}[t]
    \centering

    \includegraphics[width=0.32\linewidth]{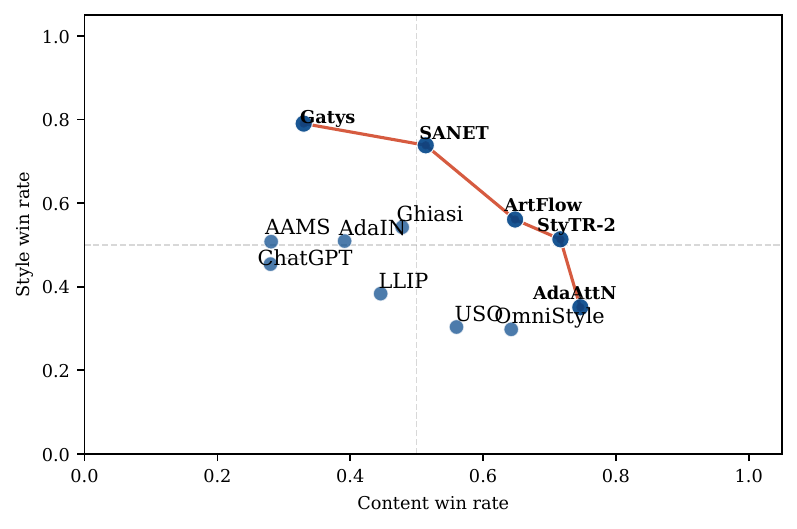}
    \includegraphics[width=0.32\linewidth]{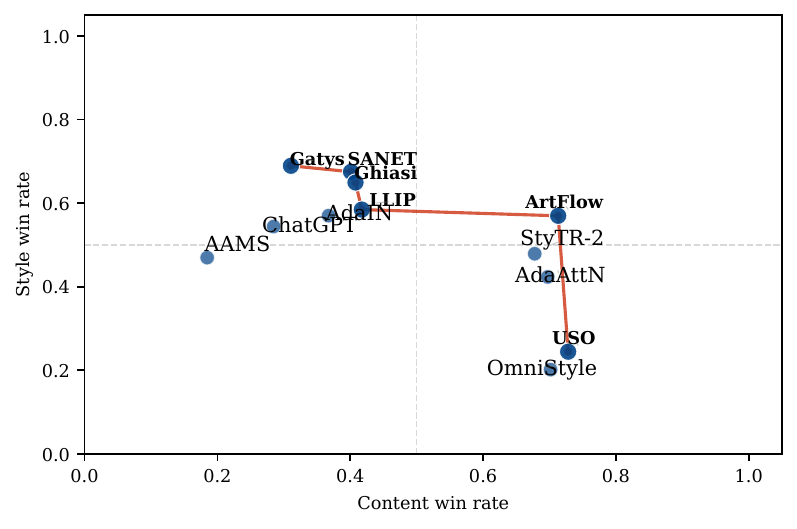}
    \includegraphics[width=0.32\linewidth]{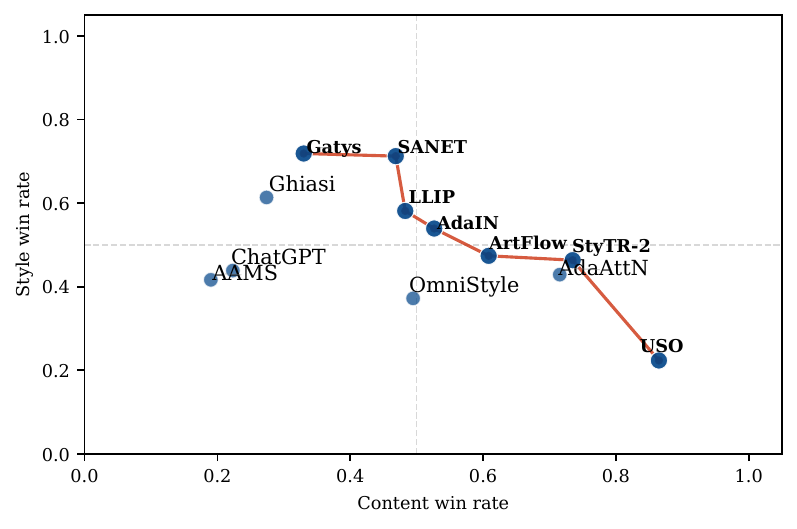}
     \vspace{4pt}
    \includegraphics[width=0.32\linewidth]{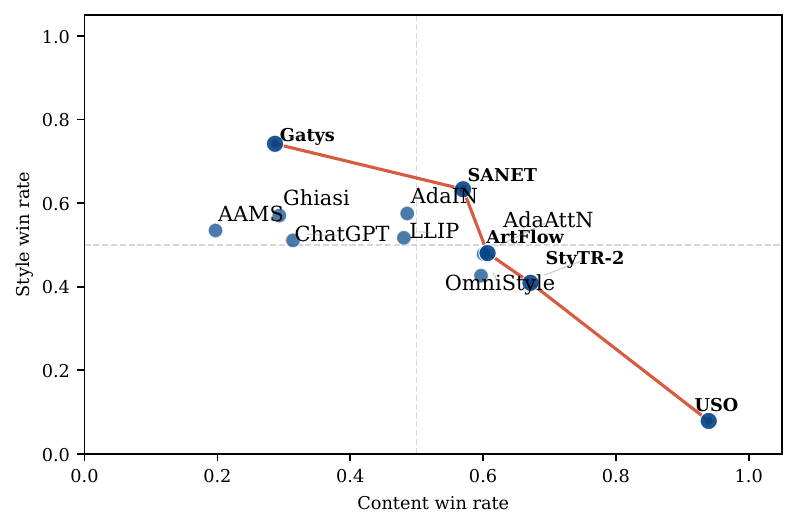}
    \includegraphics[width=0.32\linewidth]{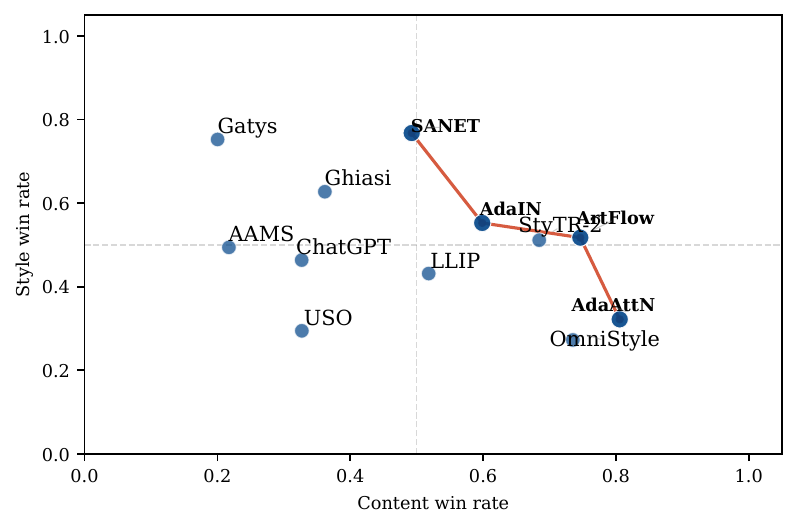}
    \includegraphics[width=0.32\linewidth]{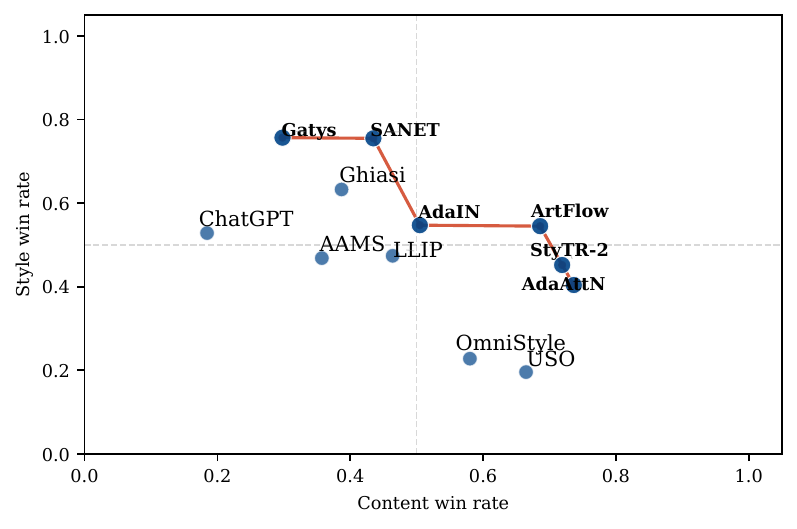}

    \caption{
    Content-style preference across different content images.
    Each subplot corresponds to a specific content image, arranged from left to right and top to bottom as:
    Angel, Athletes, Barn, Berries, Daisy, and Mac.
    Each point represents a stylisation method positioned by its content and style win rates.
    The red curve indicates the Pareto frontier of the two objectives, which highlights the optimal trade-off boundary between the two criteria.
    Compared to the per-style analysis, the distributions remain largely consistent across different contents, indicating that the trade-off is relatively stable with respect to the input image.
    }

    \label{fig:pareto_all_contents}
\end{figure*}

\subsection{Qualitative Comparison Across Methods.}
To complement the quantitative analysis, Fig.~\ref{fig:top3_bottom3_each_method} shows the top-3 and bottom-3 stylised results for each method ranked by overall human preference.

Top-ranked results generally achieve a better balance between content preservation and stylistic expression, while bottom-ranked results often exhibit either structural distortions or weak style transfer. This behaviour varies across methods: some favour strong style at the cost of content fidelity, whereas others preserve structure but lack distinctive style. These observations are consistent with the Pareto analysis, highlighting the inherent trade-off between the two objectives.

\section{Training Details of Regression Heads}
\label{sec:hyperparameters}
This section describes the implementation
and the optimisation settings used during training. These design choices aim to provide an effective learning framework that aligns predicted scores with human preference signals while maintaining stability and reproducibility.
Each regression head is implemented as a lightweight two-layer multilayer perceptron (MLP) with ReLU activations and dropout. The hidden dimension is set to 256 with a dropout rate of 0.2, and the final layer outputs a scalar score.

Each head is trained independently using human-preference-derived targets (Rank Centrality scores). The training setup is as follows:
\begin{itemize}
    \item \textbf{Loss function:} Smooth $L_1$ loss
    \item \textbf{Optimizer:} AdamW
    \item \textbf{Learning rate:} $2 \times 10^{-4}$
    \item \textbf{Weight decay:} $1 \times 10^{-4}$
    \item \textbf{Batch size:} 32
    \item \textbf{Training epochs:} 40
\end{itemize}

{\color{black}

\section{Backbone Architecture Ablation}
\label{app:backbone_ablation}

We further investigate the effect of the feature extraction backbone used in ASTRA-Score. We compare architectures from several representative backbone families, including MobileNetV3, VGG, ResNet, ViT, and DINOv2. For all variants, the pretrained backbone is kept frozen and only the task-specific regression heads are trained, such that the comparison primarily reflects the suitability of the extracted representations for modelling human perceptual judgements.

Table~\ref{tab:backbone_ablation} reports the results under the leave-one-method-out (LOMO) evaluation protocol. For each backbone, we report the mean and standard deviation of Spearman's $\rho$ and Kendall's $\tau$ across the held-out stylisation methods for content preservation, style similarity, and overall preference.

\begin{table*}[t]
\centering
\caption{Backbone ablation for ASTRA-Score under leave-one-method-out (LOMO) evaluation. Results are reported as the mean $\pm$ standard deviation of Spearman's $\rho$ and Kendall's $\tau$ across held-out stylisation methods. The best result for each criterion and correlation metric is shown in bold.}
\label{tab:backbone_ablation}

\begin{tabular}{lcccccc}
    \toprule
    & \multicolumn{2}{c}{\textbf{Content}}
    & \multicolumn{2}{c}{\textbf{Style}}
    & \multicolumn{2}{c}{\textbf{Overall}} \\
    \cmidrule(lr){2-3}
    \cmidrule(lr){4-5}
    \cmidrule(lr){6-7}
    Backbone
    & Spearman & Kendall
    & Spearman & Kendall
    & Spearman & Kendall \\
    \midrule

    ResNet18
    & 0.658 $\pm$ 0.209 & 0.489 $\pm$ 0.164
    & 0.674 $\pm$ 0.076 & 0.497 $\pm$ 0.062
    & 0.580 $\pm$ 0.156 & 0.418 $\pm$ 0.118 \\

    ResNet34
    & 0.628 $\pm$ 0.198 & 0.464 $\pm$ 0.157
    & 0.672 $\pm$ 0.077 & 0.492 $\pm$ 0.063
    & 0.566 $\pm$ 0.179 & 0.410 $\pm$ 0.135 \\

    ResNet50
    & 0.656 $\pm$ 0.173 & 0.488 $\pm$ 0.135
    & 0.707 $\pm$ 0.088 & 0.521 $\pm$ 0.074
    & 0.608 $\pm$ 0.156 & 0.442 $\pm$ 0.123 \\

    VGG19
    & 0.671 $\pm$ 0.138 & 0.491 $\pm$ 0.115
    & 0.659 $\pm$ 0.082 & 0.480 $\pm$ 0.067
    & 0.622 $\pm$ 0.119 & 0.453 $\pm$ 0.095 \\

    ViT-B/16
    & 0.691 $\pm$ 0.152 & 0.513 $\pm$ 0.127
    & 0.648 $\pm$ 0.136 & 0.475 $\pm$ 0.115
    & 0.505 $\pm$ 0.164 & 0.360 $\pm$ 0.128 \\

    DINOv2
    & 0.604 $\pm$ 0.180 & 0.439 $\pm$ 0.150
    & 0.570 $\pm$ 0.156 & 0.410 $\pm$ 0.121
    & 0.481 $\pm$ 0.254 & 0.345 $\pm$ 0.187 \\

    MobileNetV3
    & 0.686 $\pm$ 0.161 & 0.512 $\pm$ 0.136
    & 0.663 $\pm$ 0.128 & 0.490 $\pm$ 0.106
    & 0.624 $\pm$ 0.109 & 0.456 $\pm$ 0.089 \\

    VGG11
    & \textbf{0.718 $\pm$ 0.144} & \textbf{0.540 $\pm$ 0.128}
    & \textbf{0.714 $\pm$ 0.098} & \textbf{0.531 $\pm$ 0.084}
    & \textbf{0.667 $\pm$ 0.099} & \textbf{0.492 $\pm$ 0.081} \\

    \bottomrule
\end{tabular}

\end{table*}

VGG11 achieves the strongest performance across all three evaluation criteria. It obtains Spearman correlations of $0.718$, $0.714$, and $0.667$ for content preservation, style similarity, and overall preference, respectively, with corresponding Kendall correlations of $0.540$, $0.531$, and $0.492$. Compared with MobileNetV3, which was also considered as a more compact backbone, VGG11 improves the mean Spearman correlation by $0.032$, $0.051$, and $0.043$ for content, style, and overall preference, respectively.

Increasing backbone capacity does not consistently improve alignment with human judgements. In particular, VGG19 performs below VGG11 across all three criteria, while ResNet50 is competitive for style similarity but remains weaker for content preservation and overall preference. The transformer-based ViT-B/16 and DINOv2 variants likewise do not provide consistent improvements over VGG11. These results indicate that the choice of backbone has a non-negligible effect on perceptual evaluation performance, and motivate our use of VGG11 as the default feature extractor for ASTRA-Score.

}
\end{document}